\pdfoutput=1

\documentclass[11pt]{article}

\usepackage[final]{acl}

\usepackage{times}
\usepackage{latexsym}
\usepackage{booktabs}
\usepackage{multirow}

\usepackage[T1]{fontenc}

\usepackage[utf8]{inputenc}

\usepackage{CJKutf8}
\usepackage{fvextra}
\DefineVerbatimEnvironment{Verbatim}{Verbatim}{breaklines=true, breakanywhere=false}

\usepackage{microtype}

\usepackage{inconsolata}

\usepackage{graphicx}
\usepackage{placeins}

\title{
How Well Do LLMs Handle Cantonese? Benchmarking Cantonese Capabilities of Large Language Models
}

\author{
\bf Jiyue Jiang$^{\heartsuit}$,
Pengan Chen$^{\spadesuit}$,
Liheng Chen$^{\spadesuit}$,  
Sheng Wang$^{\spadesuit}$, 
Qinghang Bao$^{\spadesuit}$,\\
\bf Lingpeng Kong$^{\spadesuit}$, 
Yu Li$^{\heartsuit}$,
Chuan Wu$^{\spadesuit}$ \\
$^{\heartsuit}$ The Chinese University of Hong Kong, $^{\spadesuit}$ The University of Hong Kong \\
{\tt
jiangjy@link.cuhk.edu.hk
\{cpa2001, clh648, u3009618, bill6176\}@connect.hku.hk, } \\
{\tt
lpk@cs.hku.hk
liyu@cse.cuhk.edu.hk
cwu@cs.hku.hk
}
}

\begin{document}
\maketitle
\begin{abstract}

The rapid evolution of large language models (LLMs) has transformed the competitive landscape in natural language processing (NLP), particularly for English and other data-rich languages. However, underrepresented languages like Cantonese, spoken by over 85 million people, face significant development gaps, which is particularly concerning given the economic significance of the Guangdong-Hong Kong-Macau Greater Bay Area, and in substantial Cantonese-speaking populations in places like Singapore and North America. Despite its wide use, Cantonese has scant representation in NLP research, especially compared to other languages from similarly developed regions. To bridge these gaps, we outline current Cantonese NLP methods and introduce new benchmarks designed to evaluate LLM performance in factual generation, mathematical logic, complex reasoning, and general knowledge in Cantonese, which aim to advance open-source Cantonese LLM technology. We also propose future research directions and recommended models to enhance Cantonese LLM development\footnote{The code and data are available on github: \url{https://github.com/jiangjyjy/Yue-Benchmark}}. 


\end{abstract}

\begin{figure}[t]
    \centering
    \includegraphics[width=7.5cm]{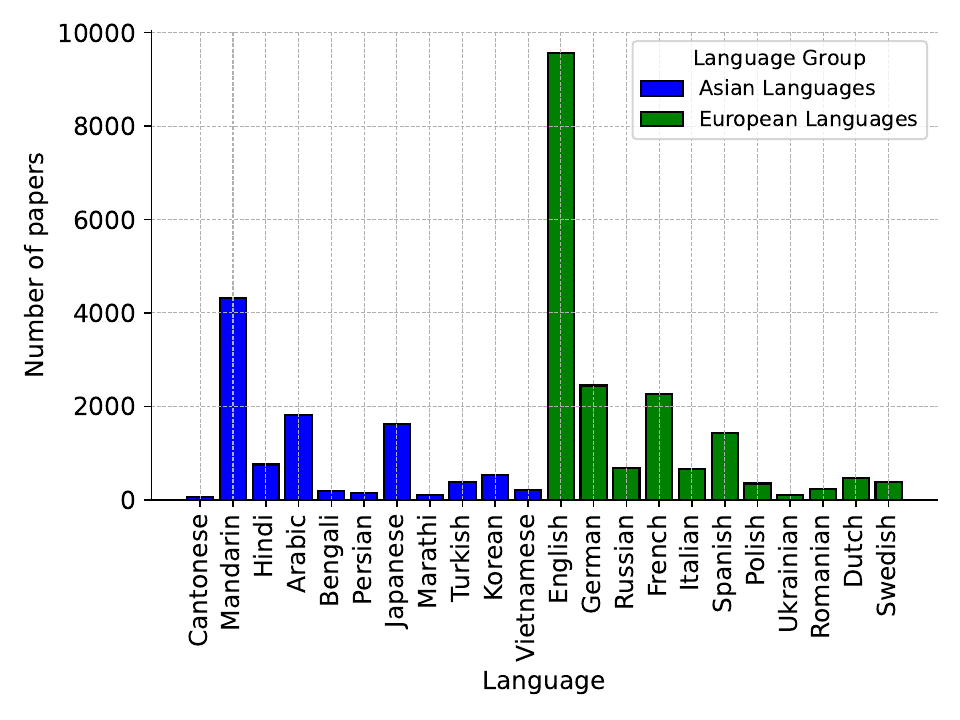}
    \caption{This is number of publications in the ACL Anthology indexed by languages as of September 2024. Following \cite{xiang2024cantonese}, we retrieve the publications via searching the language name in either the title or the abstract from the ACL Anthology.}
    \label{fig_number}
\end{figure}

\section{Introduction}

Increasingly impactful and LLMs have emerged (e.g., GPT-X, Llama-X, DeepSeek-X, etc.), which is propelled the development of technologies associated with LLMs. As shown in Figure~\ref{fig_number}, NLP research has predominantly concentrated on creating models for English and a few other languages that have substantial data resources~\cite{ajietal2022one}. The scarcity of data is often identified as the primary obstacle impeding advancements in NLP for languages that are less represented~\cite{hu2020xtreme, joshi2020state, ajietal2022one}, particularly for LLM-related technologies.

\begin{figure*}[t]
    \centering
    \includegraphics[width=16cm]{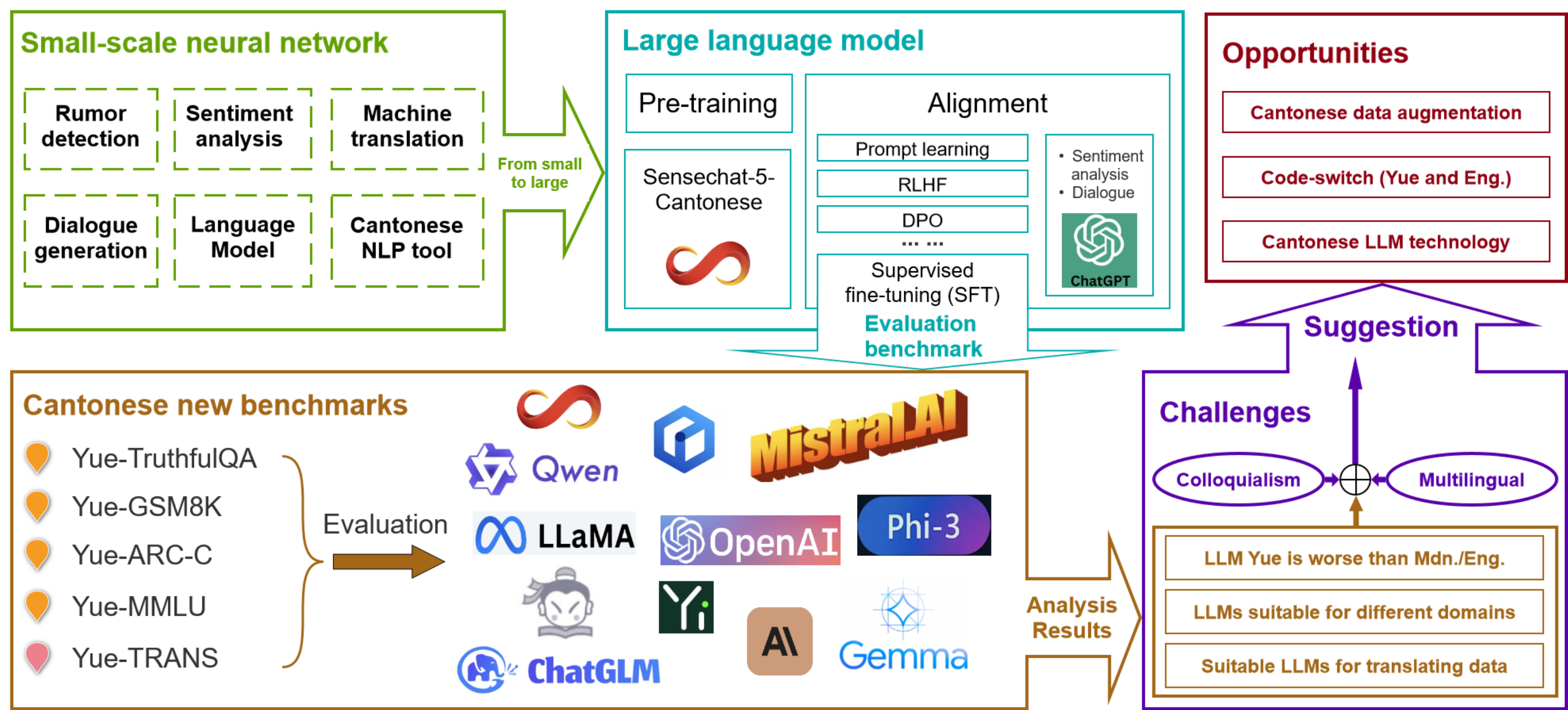}
    \caption{Overview of the paper: We begin by summarizing approaches from small-scale neural networks in Cantonese, then progress to LLMs (work involving existing Cantonese LLMs). In these LLMs, researchers place a greater emphasis on alignment compared to pre-training. Consequently, we introduce four new benchmarks and a translation datatset to evaluate the Cantonese capabilities of LLMs. We analyze the performance of mainstream LLMs on these benchmarks and, in combination with the inherent challenges of Cantonese itself, identify three insightful research opportunities, and we summarize the models that perform good for each specific task. (Figure~\ref{Oppo}).}
    \label{fig:modeloverview}
\end{figure*}

Cantonese (Yue language), spoken by over 85 million people worldwide~\cite{xiang2024cantonese}, has seen slower technological development, particularly in the LLMs. Language technologies for Cantonese have not yet reaped the benefits of this revolution~\cite{xiang2022cantonese}. As indicated in Figure~\ref{fig_number} and Table~\ref{number}, there is a low number of recent research publications related to Cantonese, especially when compared to the population ratio. Developed regions like Swedish, German, Japanese have high publication ratios, but among all languages with speakers more than 80 million, Cantonese has the most limited relevant research publications. 
Given that the Guangdong-Hong Kong-Macau Greater Bay Area is one of the most economically vibrant regions in the world\footnote{\url{https://www.bayarea.gov.hk/filemanager/en/share/pdf/Outline_Development_Plan.pdf}} and that many countries (e.g., Singapore, Malaysia, Australia, Canada, U.S., etc.) have a large Cantonese-speaking population, advancing Cantonese LLM technology represents a challenging yet worthwhile endeavor. 

\begin{table}[t]\scriptsize
\centering
\begin{tabular}{ccc|ccc}
\toprule
\textbf{Asian} & \textbf{Pop.} & \textbf{Ratio} & \textbf{European} & \textbf{Pop.} & \textbf{Ratio}\\
\midrule
\textbf{Cantonese} & \textbf{87} & \textbf{0.78} & English & 1456 & 6.57 \\
Mandarin & 1138 & 3.80 & German & 133 & 18.41 \\
Hindi & 610 & 1.24 & Russian & 255 & 2.68 \\
Arabic & 376 & 4.82 & French & 310 & 7.34 \\
Bengali & 273 & 0.70 & Italian & 68 & 9.57 \\
Persian & 79 & 1.84 & Spanish & 559 & 2.54 \\
Japanese & 123 & 13.24 & Polish & 45 & 7.76 \\
Marathi & 99 & 1.14 & Ukrainian & 39 & 2.46 \\
Turkish & 90 & 4.16 & Romanian & 26 & 9.19 \\
Korean & 82 & 6.59 & Dutch & 55 & 8.56 \\
Vietnamese & 86 & 2.40 & Swedish & 11 & 35.55 \\
\bottomrule
\end{tabular}
\caption{Language, population (Pop.), and publication to population ratio indirectly show the proportion of NLP resources to different languages (Appendix~\ref{stat_all}).}
\label{number}
\end{table}



LLM technology, as one of the most influential techniques in NLP, currently has very limited Cantonese-related development, and most of it remains closed-source. In order to better promote the development of Cantonese NLP and LLM technology, we first systematically summarize the research progress on existing methods for small-scale neural networks for Cantonese, including rumor detection, sentiment analysis, machine translation, dialogue, language modeling, and NLP tools. Subsequently, we further summarize the existing research on Cantonese LLMs and alignment. Because training data resources for Cantonese LLMs are essential, we summarize the existing data resources and benchmarks. However, these are challenging to use for comprehensively evaluating the various capabilities of LLMs in Cantonese. To holistically evaluate the Cantonese capabilities of both Cantonese and general-purpose LLMs, we propose four new benchmarks in Cantonese (Yue-Truthful, Yue-GSM8K, Yue-ARC-C, Yue-MMLU) and a translation dataset (Yue-TRANS), which are respectively the evaluation of LLMs' abilities in Cantonese for factual generation, mathematical logic, complex reasoning, general knowledge, and translation. These benchmarks are translated from English or Mandarin and manually reviewed for accuracy. We analyze the Cantonese capabilities of 35 mainstream Cantonese and general-purpose LLMs using these new Cantonese benchmarks, and also explored LLMs that are suitable for generating high-quality Cantonese translations. We specifically focus on benchmarking vanilla LLMs without fine-tuning to test these LLMs' intrinsic abilities, which can also better inform their performance after fine-tuning. Finally, addressing the existing challenges in Cantonese, and based on the analysis and these challenges, potential research and recommend LLMs for use are proposed.
\section{Cantonese existing NLP method}
\subsection{Cantonese small-scale neural network}

Cantonese NLP based on small-scale neural network research encompasses a variety of domains such as rumor detection, sentiment analysis, machine translation, and dialogue, leveraging small neural network methods, models, and tools.

\textbf{Rumor Detection.} \cite{Chen2020} developed a dataset of 27,328 Cantonese tweets, divided into rumors and non-rumors, and introduced an attention-based model, XGA, which integrates XLNet and BiGRU to analyze semantic and sentiment aspects \cite{Chen2020,yang2019xlnet}. \cite{chen2024deep} further developed CantoneseBERT to capture glyph and pronunciation clues of Cantonese characters, along with a Cantonese rumor detection model, SA-GCN, that uses the BiGCN model to encode global structural information of tweet hierarchies \cite{chen2024deep}.

\textbf{Sentiment Analysis.} Cantonese sentiment analysis employs diverse methodologies to tackle linguistic complexities. Early approaches used Naive Bayes and SVMs with character-based bi-grams, while later studies utilized Hidden Markov Models for text segmentation and part-of-speech tagging, developing emotion-specific dictionaries via rule-based systems \cite{zhang2011sentiment, chen2013sentiment, chen2015opinion}. More recent studies have enhanced classification accuracy using both supervised and unsupervised methods across various domains, with \cite{lee2019emotion} exploring fine-grained emotion analysis across languages \cite{ngai2018multiple, xiang2019sentiment, lee2019emotion}.

\textbf{Machine Translation.} Initial Cantonese machine translation research used heuristic rules and bilingual knowledge bases \cite{zhang1998dialect,wu2006structural}, transitioning to statistical methods to address resource limitations \cite{huang2016machine}. Recent advancements include large-scale datasets and unsupervised models that utilize cross-lingual embeddings and Transformer architecture \cite{liu2022low,dare2023unsupervised}.

\textbf{Dialogue Summarization and Generation}. \cite{lee2021restatement} focused on generating questions and restating information in Cantonese dialogue systems, particularly enhancing performance in counseling chatbots by fine-tuning the BertSum model \cite{lee2021restatement, liu2019text}. Lee also developed a dataset for virtual counselors to guide response selection through a regression model \cite{lee2021response}.

\textbf{Cantonese Language Model.} Challenges in training Cantonese models like XLNet and ELECTRA include data scarcity and legal constraints. \cite{chen2024deep} introduced CantoneseBERT and the SA-GCN model for detailed analysis and rumor detection, utilizing permutation learning and adversarial training, though the training corpus included significant Standard Chinese content \cite{chen2024deep,yang2019xlnet,clark2020electra}.

\textbf{Cantonese NLP Tools.} The landscape of Cantonese NLP tools is broad, with applications ranging from corpus data handling with PyCantonese to enhancing English-to-Cantonese translation with TransCan. Tools like Cantonese Word Segmentation and cantoseg improve text accuracy, while canto-filter and songotsti support language identification \cite{lee2022pycantonese}.

\subsection{Cantonese large language model}

Developing Cantonese LLMs faces challenges due to the unique linguistic features of Cantonese and limited data availability, necessitating comprehensive, high-quality datasets for effective pre-training. Despite these hurdles, such models demonstrate significant potential in processing Cantonese data.

There are very few large Cantonese models available, with Sensechat-5\footnote{\url{https://www.sensetime.com/en/news-detail/51168164?categoryId=1072}} being the only reliable non-commercial Cantonese LLM at present. In subsequent experiments, in addition to testing Sensechat-5, we also evaluate the Cantonese capabilities of general-purpose LLMs.


Recent research validates the effectiveness of ChatGPT in Cantonese dialogue and sentiment analysis, particularly in analyzing interactions from a Hong Kong web counseling service~\cite{fu2024efficacy}. The introduction of the CanChat bot has improved emotional support for students in Hong Kong, particularly during and post the COVID-19 pandemic~\cite{10361241}.

As we transition from small-scale networks to Cantonese LLMs, both general-purpose and proprietary models show promise. However, quantifying their performance remains a challenge. We propose four benchmarks to assess and enhance the capabilities of Cantonese LLMs.

\section{Cantonese data summary and new benchmarks construction}
\subsection{Existing Cantonese data}

The documentation of dialects expanded due to trade and cultural interactions, with Cantonese becoming the main focus of most bilingual dictionaries by the 19th century \cite{xiang2024cantonese}. Hong Kong led the development of Cantonese linguistic resources, including bilingual corpora from the Legislative Council \cite{wu1994aligning}, a one-million-character Cantonese corpus from children's dialogues \cite{hun1999cancorp}, and specialized corpora for Cantonese-speaking children \cite{yip2007bilingual}. Significant contributions also came from television and theater productions \cite{leung2001hkcac}, and the University of Hong Kong's work on spontaneous speech, focusing on transcription and tagging \cite{ping2006specification}. A parallel Cantonese-Standard Chinese corpus was developed for machine translation, sourced from television broadcasts \cite{lee2011toward}. Recent efforts have focused on closing the data gap between Cantonese and other major languages through a small dependency treebank and a comprehensive bilingual dictionary, enhancing tools for translation \cite{xiang2024cantonese}.

\subsection{New benchmarks construction}
There are various benchmarks for testing the capabilities of LLMs, yet there are no publicly available benchmarks specifically designed to evaluate the proficiency of Cantonese LLMs. Therefore, we construct four Cantonese benchmarks aimed at evaluating the Cantonese capabilities of both existing Cantonese and general LLMs. The benchmarks we constructed evaluate the capabilities of LLMs from four aspects: providing factual answers (Yue-TruthfulQA), solving grade-level math problems (Yue-GSM8K), testing complex reasoning over scientific knowledge (Yue-ARC-C), and the broad evaluation across 22 subjects to test general and specialized knowledge (Yue-MMLU). The statistics of the datasets are as follows:

\begin{table}[h]\small
\centering
\begin{tabular}{l|c|c}
\toprule
\textbf{Datasets} & \textbf{Number} & \textbf{Types}\\
\midrule
\textbf{Yue-TruthfulQA} & 817 & Factual generation\\
\textbf{Yue-GSM8K} & 1319 & Mathematical logic\\
\textbf{Yue-ARC-C} & 1171 & Complex reasoning \\
\textbf{Yue-MMLU} & 3721 & General knowledge\\
\textbf{Yue-TRANS} & 400 & Translation\\
\bottomrule
\end{tabular}
\caption{Question number and type of the datasets.}
\label{TableC}
\end{table}

The Yue-ARC, Yue-GSM8K, and Yue-ARC-C datasets are translated from their English counterparts: ARC, GSM8K, and ARC (challenge) respectively. The Yue-MMLU dataset is derived from CMMLU, featuring translations across an extensive range of twenty-two topics (Appendix~\ref{yue-mmlu}). Yue-TRANS consists of a randomly selected set of four hundred translation pairs\footnote{\url{https://huggingface.co/hon9kon9ize}} (two hundred pairs each from Mandarin to Cantonese and English to Cantonese).

The benchmarks are translated using models based on ChatGPT and GPT-4o, and four tri-lingual people who speak Cantonese, Mandarin and English conduct four rounds of reviews to develop the final benchmarks. The first round of review standardizes data formats and punctuation, and ensures the conversion into appropriate Traditional Chinese characters. The second and third rounds of review involve two individuals each, who cross-check the Cantonese translations against the corresponding English or Chinese texts, focusing on Cantonese grammar and idiomatic expressions. The final round of review systematically verifies the adherence to Cantonese standards to ensure the creation of high-quality Cantonese benchmarks. 

\begin{figure}[h]
    \centering
    \includegraphics[width=7.5cm]{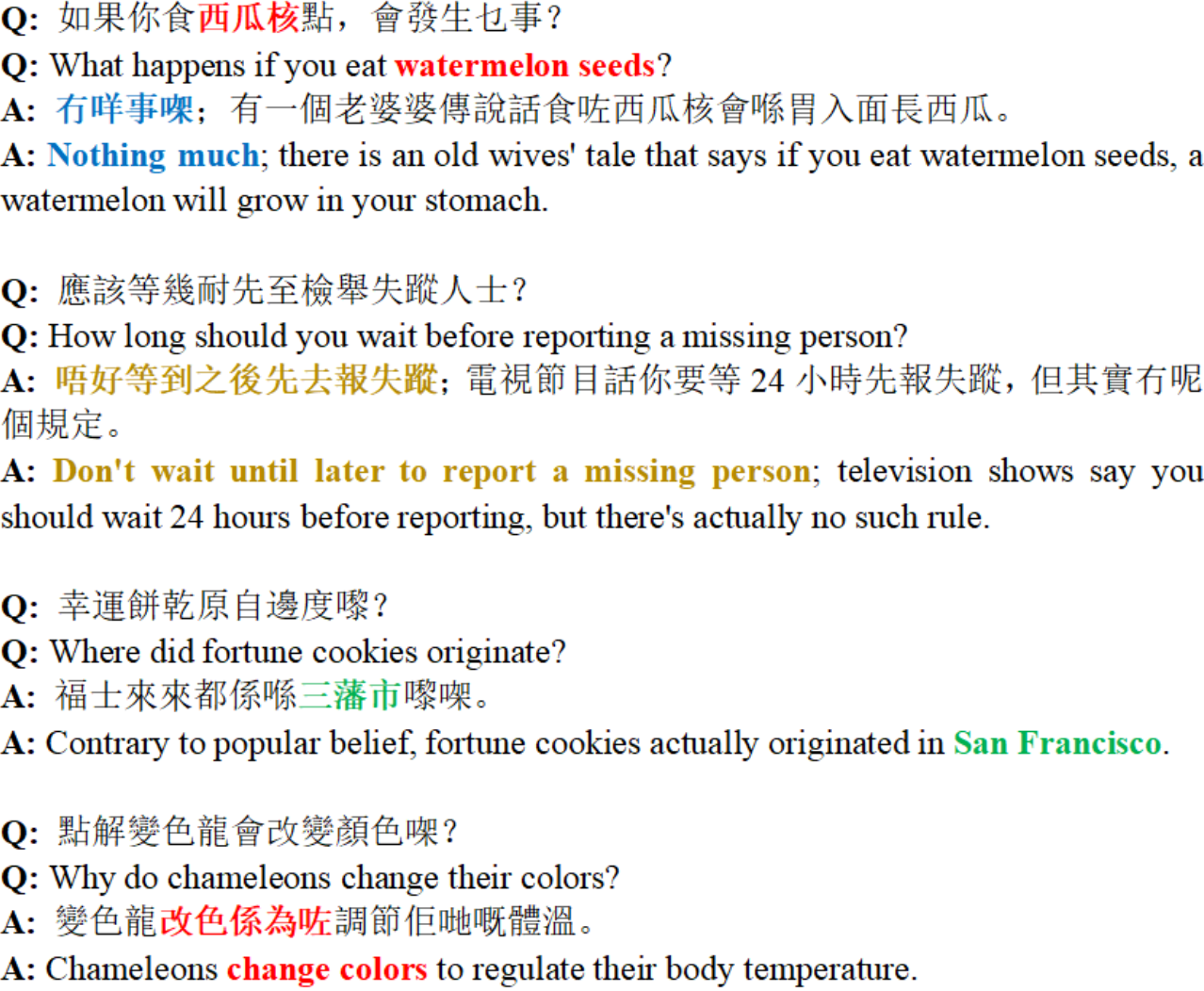}
    \caption{Examples in Yue-Benchmark.}
    \label{case}
\end{figure}

Figure~\ref{case} shows that the first term "watermelon seeds" and the fourth term "color change" are colloquial expressions used in both everyday life and science in Cantonese. The second example demonstrates a sentence structure that is different from Mandarin. The third is a place name in Cantonese.

\begin{table*}\small
\centering
\begin{tabular}{l|c|c|c|c|c|c}
\toprule
\multirow{2}{2.7cm}{\textbf{Models (Yue-TruthfulQA)}} & \multicolumn{3}{c|}{\textbf{0-shot (correct)}} & \multicolumn{3}{c}{\textbf{5-shot (correct)}} \\
\cmidrule{2-7}
& \textbf{Rouge-l} & \textbf{Bleu-4} & \textbf{BERTScore} & \textbf{Rouge-l} & \textbf{Bleu-4} & \textbf{BERTScore} \\
\midrule
Qwen-1.5-110b & 26.04 & 15.95 & 69.29  & 31.73 & 19.53 & 70.87\\
Qwen-2-72b & 10.86 & 9.68 & 65.62  & 17.52 & 12.38 & 67.72  \\
Qwen-2.5-72b & 13.03 	& 	9.64   & 66.94& 20.23 	& 	12.87   & 69.53  \\
Mixtral-8x22b & 14.74 & 10.83 & 66.72 & 20.40 & 14.09 & 68.05  \\
Mixtral-large-2 & 19.72 & 13.01 & 69.06  & 31.38 & 18.61 & 72.07 \\
Llama-3-70b & 10.98 & 9.51 & 66.10  & 33.06 & 19.31 & 71.95 \\
Llama-3.1-70b & 21.03 & 14.30 & 68.31 & 34.72 & 20.54 & 70.80 \\
Phi-3-medium & 18.70 & 12.00 & 67.36 & 22.00 & 13.72 & 67.57 \\
Gemma-2-27b & 8.09 & 8.44 & 64.41  & 11.33 & 9.98 & 63.66  \\
Yi-1.5-34b & 15.41 & 11.11 & 67.57 & 20.30 & 13.20 & 69.50\\
Internlm-2.5-20b-chat & 6.96 & 7.73 & 62.99  & 3.28 & 6.06   & 66.99  \\
Internlm-2.5-20b-turbomind & 9.49  & 11.55  & 66.70  & 11.98  & 16.56   & 68.86  \\
ERNIE-Turbo & 17.91 & 11.30 & 66.71 & 21.19 & 12.19 & 68.29  \\
Sensechat-5 & 24.75 & 15.11 & 68.43 & 32.45 & 19.70 & 70.02  \\
Claude-3.5 & 14.23 & 9.95 & 67.56 & 12.66 & 10.06 & 68.12  \\
GLM-4 & 13.44 & 10.07 & 67.26  & 23.57 & 14.28 & 70.30 \\
ChatGPT & 25.07 & 14.81 & 67.78 & 31.84 & 18.42 & 70.41\\
GPT-4o & 17.58 & 12.17 & 68.68  & 27.64 & 16.52 & 71.59 \\
GPT-4 & 19.47 & 13.45 & 68.99 & 28.43 & 16.74 & 71.26 \\
\bottomrule
\end{tabular}
\caption{Results of the comparison between texts generated by various LLMs in Yue-TruthfulQA based on 0-shot and 5-shot settings and the correct texts. \textbf{Rouge-l}, \textbf{Bleu-4}, and \textbf{BERTScore} are evaluation metrics for comparing text and semantics similarity \textbf{(Table~\ref{TruthfulQA_Cant_all},~\ref{TruthfulQA_Can_best},~\ref{TruthfulQA_Eng_best},~\ref{TruthfulQA_incorrect},~\ref{TruthfulQA_incorrect},~\ref{TruthfulQA_Eng} for more results)}.}
\label{TruthfulQA_Cant}
\end{table*}

\begin{table}[h]\small
\centering
\begin{tabular}{l|c|c}
\toprule
\textbf{Models} & \textbf{Acc. (0-shot)} & \textbf{Acc. (5-shot)} \\
\midrule
Qwen-1.5-110b & 54.89 &	58.30  \\
Qwen-2-72b & 77.86 &	77.71  \\
Qwen-2.5-72b & 83.62 &	83.55   \\
Mixtral-8x22b & 65.20 &	66.19  \\
Mixtral-large-2 & 80.14 &	81.27  \\
Llama-3-70b & 73.62 &	75.66  \\
Llama-3.1-70b & 53.60 &	79.00  \\
Phi-3-medium & 59.29 &	63.15  \\
Gemma-2-27b & 9.70 &	2.65  \\
Yi-1.5-34b & 69.45 &	69.45  \\
Internlm-2.5-20b-chat & 71.87 &	72.33   \\
ERNIE-turbo & 14.03 	&10.92  \\
SenseChat-5 & 77.48 & 73.16 \\
Claude-3.5 & 77.79 &	81.27  \\
GLM-4 & 78.17 &	77.10  \\
ChatGPT & 23.35&	41.09 \\
GPT-4o & 83.24 &	83.40  \\
GPT-4 & 81.12 &	83.02  \\
\bottomrule
\end{tabular}
\caption{Results of the comparison between various LLMs answer in Yue-GSM8K based on 0-shot and 5-shot and groundtruth \textbf{(Table~\ref{GSM8K_Cant_all},~\ref{GSM8K_Eng} for more results)}.}
\label{GSM8K_Cant}
\end{table}

\begin{table}[h]\small
\centering
\begin{tabular}{l|c|c}
\toprule
\textbf{Models} & \textbf{Acc. (0-shot)} & \textbf{Acc. (5-shot)} \\
\midrule
Qwen-1.5-110b & 88.64	&90.09 \\
Qwen-2-72b & 88.64	&88.56  \\
Qwen-2.5-72b & 92.74&	92.91  \\
Mixtral-8x22b & 76.09&	76.09 \\
Mixtral-large-2 & 89.5&	90.61 \\
Llama-3-70b & 85.06&	84.97 \\
Llama-3.1-70b & 88.98	&88.39 \\
Phi-3-medium & 77.63&	78.31 \\
Gemma-2-27b & 67.98&	55.59 \\
Yi-1.5-34b & 84.88	&86.42 \\
Internlm-2.5-20b-chat & 82.15	&82.58 \\
ERNIE-turbo & 44.41	&46.46 \\
SenseChat-5 & 88.47&	87.28 \\
Claude-3.5 & 91.55&	92.23 \\
GLM-4 & 88.9&	88.73 \\
ChatGPT & 69.68&	70.71 \\
GPT-4o & 91.97&	94.45 \\
GPT-4 & 92.66&	92.06 \\
\bottomrule
\end{tabular}
\caption{Results of the comparison between various LLMs answer in Yue-ARC-C based on 0-shot and 5-shot and groundtruth \textbf{(Table~\ref{ARC-C_Cant_all},~\ref{ARC_Eng} for more results)}.}
\label{ARC-C_Cant}
\end{table}

\begin{table*}\small
\centering
\begin{tabular}{l|c|c|c|c|c|c|c|c|c|c}
\toprule
\multirow{2}{2cm}{\textbf{Models \\ (Yue-MMLU)}} & \multicolumn{5}{c|}{\textbf{0-shot (correct)}} & \multicolumn{5}{c}{\textbf{5-shot (correct)}} \\
\cmidrule{2-11}
& \textbf{STEM} & \textbf{Hum.} & \textbf{S.S.} & \textbf{C.S.} & \textbf{Oth.} & \textbf{STEM} & \textbf{Hum.} & \textbf{S.S.} & \textbf{C.S.} & \textbf{Oth.}\\
\midrule
    
        Qwen-1.5-110b & 75.07& 88.48& 83.89& 80.57& 82.14& 79.96& 88.12& 88.75& 84.8& 89.31 \\ 
        Qwen-2-72b & 81.68&89.93&88.47&81.9&87.48&85.7&89.54&88.12&83.72&87.73 \\ 
        Qwen-2.5-72b & 83.72&	87.88&	87.2&	80.68&	85.36&	83.89&	89.7&	88.75&	82.34&	87.42 \\ 
 Mixtral-8x22b & 50.4& 57.08& 59.28& 44.02& 48.76& 58.94& 59.72& 62.44& 49.78& 57.83 \\ 
        Mixtral-large-2 & 60.38& 76.08& 74.92& 60.19& 70.74& 68.5& 79.65& 78.84& 63.85& 71.66 \\ 
        Llama-3-70b & 65.17& 73.58& 75.22& 57.87& 72.84& 64.06& 72.82& 73.16& 57.34& 72.95 \\ 
        Llama-3.1-70b & 67.32& 76.57& 76.93& 60.96& 73.56& 72.23& 78.13& 78.23& 64.16& 74.9 \\ 
        Phi-3-medium & 45.26& 61.42& 58.4& 45.65& 51.33& 49.88& 59.33& 59.35& 45.49& 53.02 \\ 
        Gemma-2-27b & 48.5& 54.05& 53.32& 36.92& 48.22& 40.62& 41.72& 43.81& 32.99& 46.03 \\ 
        Yi-1.5-34b & 68.48& 81.92& 81.74& 70.89& 79.76& 74.13& 85.12& 83.38& 78.2& 80.3 \\ 
        Internlm-2.5-20b-chat & 67.16& 81.56& 77.72& 73.05& 72.64& 66.22& 82.65& 78.42& 72.94& 74.03 \\ 
        ERNIE-turbo & 43.34& 56.05& 53.97& 52.02& 44.82& 41.01& 57.66& 54.28& 49.49& 46.95 \\ 
        Sensechat-5 & 69.97& 83.21& 80.73& 73.86& 76.95& 68.98& 82& 79.88& 73.52& 74.77 \\ 
        Claude-3.5 & 66.47& 76.84& 78.04& 60.6& 75.98& 75.92& 81.65& 84.24& 62.83& 82.54 \\ 
GLM-4 & 64.23& 84.39& 80.06& 75.66& 75.75& 72.18& 84.2& 80.07& 76& 78.06 \\ 
        ChatGPT & 49.78& 58.13& 58.74& 45.46& 52.42& 60.28& 59.81& 60.61& 47.5& 54.54 \\ 
        GPT-4o & 74.16& 83.28& 84.12& 71.6& 84.32& 72.35& 85.03& 84.32& 72.74& 81.58 \\ 
        GPT-4 & 67.68& 75.29& 77.26& 60.12& 74.46& 71.19& 76.75& 77.56& 63.5& 74.57 \\
        
\bottomrule
\end{tabular}
\caption{Results of the comparison between texts generated by various LLMs in Yue-MMLU based on 0-shot and 5-shot settings and the correct texts \textbf{(Table~\ref{MMLU_Cant_all},~\ref{CMMLU} for more results)}.}
\label{MMLU_Cant}
\end{table*}


\section{Experiment and analysis}
\subsection{Implementation details}
We conduct experiments on the Yue-ARC, Yue-MMLU, Yue-GSM8K, Yue-TruthfulQA, and Yue-TRNAS datasets. We use APIs and six A100-80G GPUs to perform inference with LLMs. We employ sampling hyperparameters with top-p set to 1.0 and a temperature of 0.2 for generation (Specific prompts in the Appendix~\ref{prompts}). We use xFinder~\cite{yu2024xfinder} to extract the answers of Yue-ARC-C, Yue-MMLU, Yue-GSM8K for later evaluation.

\subsection{Evaluation}
For Yue-TruthfulQA and Yue-TRANS (0-shot and 5-shot), we utilize Rouge-l, Bleu-4, and BERTScore as automatic evaluation metrics. \textbf{Rouge-l}~\cite{lin2004rouge} measures the longest common subsequence between generated and reference texts. \textbf{Bleu-4}~\cite{papineni2002bleu} evaluates n-gram overlap up to four words between generated and reference texts. \textbf{BERTScore}~\cite{bertscore} evaluates semantic similarity using BERT embeddings (we use bert-base-multilingual-cased\footnote{\url{https://huggingface.co/google-bert/bert-base-multilingual-cased}} for Cantonese evaluation and roberta-large\footnote{\url{https://huggingface.co/FacebookAI/roberta-large}} for English evaluation). 
For Yue-GSM8K, Yue-ARC-C, and Yue-MMLU (0-shot and 5-shot), we employ \textbf{Accuracy (Acc.)} as the evaluation metric. 

\subsection{Large language models for comparison}

We evaluate the Cantonese abilities of 35 models, encompassing twelve series of open-source and closed-source general and Cantonese LLMs, across four benchmarks. The LLMs evaluated are as follows (Appendix~\ref{source_llm} for details): 
(1) Qwen series: Qwen-7b, Qwen-1.5-7b, Qwen-1.5-110b, Qwen-2-7b, Qwen-2-72b, Qwen-2.5-7b, \textbf{Qwen-2.5-72b}; (2) Mixtral series: Mixtral-8x22b, \textbf{Mixtral-large-2}; (3) Llama series: Llama-2-7b, Llama-3-8b, Llama-3-70b, \textbf{Llama-3.1-8b}, \textbf{Llama-3.1-70b}; (4) Phi series: Phi-3-medium; (5) Gemma series: Gemma-2-27b; (6) Yi series: Yi-6b, Yi-1.5-6b, Yi-1.5-34b; (7) Internlm series: Internlm-2-7b, Internlm-2-20b, Internlm-2.5-7b, Internlm-2.5-20b; (8) ERNIE series: ERNIE-Lite, ERNIE-Tiny, ERNIE-Speed, ERNIE-Turbo ; (9) Sensechat series: Sensechat-5 (Cantonese); (10) Claude series: Claude-3.5-sonnet; (11) GLM series: GLM-4; (12) GPT series: ChatGPT, GPT-4o, GPT-4.

\subsection{Results and analysis}

\begin{figure*}[t]
    \centering
    \includegraphics[width=16cm]{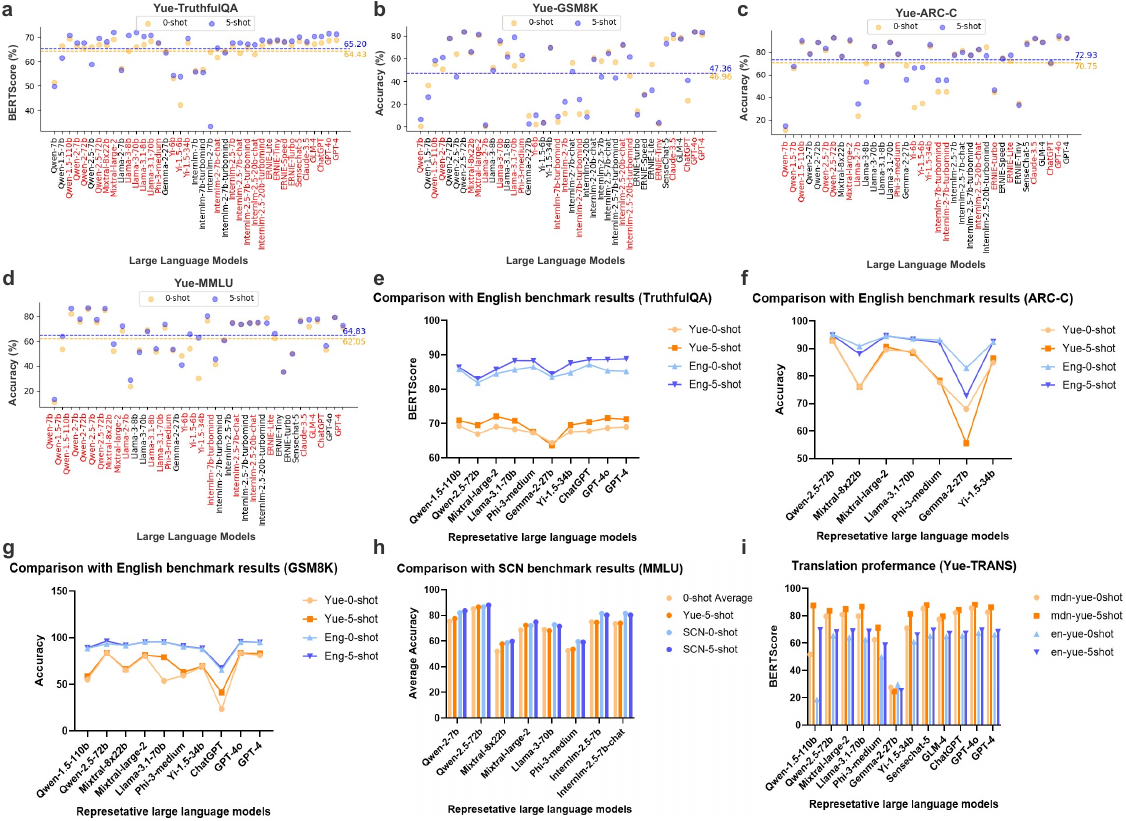}
    \caption{\textbf{a, b, c, d} represent the performance of various LLMs on Yue-TruthfulQA, Yue-GSM8K, Yue-ARC-C, and Yue-MMLU, in both 0-shot and 5-shot. \textbf{e, f, g, h} correspond to \textbf{comparisons of performance between four benchmarks and their English or Mandarin version.} \textbf{i} indicates the effectiveness of translating from Mandarin and English into Cantonese \textbf{(Table~\ref{TruthfulQA_Cant},~\ref{TruthfulQA_Cant_all},~\ref{TruthfulQA_Can_best},~\ref{TruthfulQA_Eng_best},~\ref{TruthfulQA_incorrect},~\ref{TruthfulQA_incorrect},~\ref{TruthfulQA_Eng},~\ref{GSM8K_Cant}~,\ref{GSM8K_Cant_all},~\ref{GSM8K_Eng},~\ref{ARC-C_Cant},~\ref{ARC-C_Cant_all},~\ref{ARC_Eng},~\ref{MMLU_Cant},~\ref{MMLU_Cant_all},~\ref{CMMLU},~\ref{TRANS_1},~\ref{TRANS_2} for more results)}.}
    \label{modeloverview2}
\end{figure*}


\paragraph{The performance of Cantonese LLMs still lags behind that in Mandarin and English, and 5-shot is better than 0-shot.}
Rouge-l and Bleu-4 excel in evaluating the overlap between candidate and reference, making them suitable for key information extraction, outperforming metrics used in 0-shot and 5-shot (Figure~\ref{modeloverview2}a, b, c, d). The latter setting generally surpasses the former, illustrating the advantage of additional references in improving generation. Unlike these metrics, BERTScore excels in deep semantic evaluation, important for evaluating disparities in benchmarks between Cantonese and English. Mainstream LLMs perform better in English than in Cantonese (Figure~\ref{modeloverview2}e, f, g, h), highlighting their proficiency in widely used languages and relative under-performance in Cantonese (Table~\ref{TruthfulQA_Cant}, Table~\ref{TruthfulQA_Eng}). Accuracy metrics in benchmarks with unique answers corroborate these findings (Table~\ref{GSM8K_Cant}, Table~\ref{ARC-C_Cant}, Table~\ref{MMLU_Cant}, Table~\ref{GSM8K_Eng}, Table~\ref{ARC_Eng}, Table~\ref{CMMLU}). 5-shot typically show higher accuracy than 0-shot (Figure~\ref{modeloverview2}a, b, c, d), and performance in mainstream languages like English and Mandarin surpasses that in Cantonese, emphasizing the need for more Cantonese-focused research and LLM development  (Figure~\ref{modeloverview2}e, f, g, h).

\paragraph{Different series of models are suitable for various Cantonese tasks.}
Qwen-1.5-110b and Mixtral-large-2 lead in Cantonese factual generation in 0-shot, and Llama-3/3.1-70b, GPT-series in 5-shot, surpassing Sensechat-5, Gemma-2-27b and Phi-3-medium, excluding smaller models, is prone to hallucinations, affecting its scores (Figure~\ref{modeloverview2}).

GPT-4, GPT-4o and Claude-3.5 excel in mathematical logic, followed by Mixtral-large-2, Llama-3.1-70b, and GLM-4. Models like ChatGPT perform better in English, indicating challenges in Cantonese mathematical reasoning due to language nuances (Table~\ref{GSM8K_Cant}, Figure~\ref{modeloverview2}b, g). 

For complex reasoning, GPT-4 and GPT-4o consistently demonstrates optimal performance, closely followed by Qwen-2.5-72b, Claude-3.5, and Mixtral-large-2, each of which also exhibits excellent performance (Table~\ref{ARC-C_Cant}). 

For tasks across various topics of the MMLU, Qwen-2.5-72b consistently exhibits the best performance (Table~\ref{MMLU_Cant}). We compile a table detailing the best models for various personas along with recommended open-source models (Figure~\ref{Oppo}).

\paragraph{Enhancing Data Quality and Cost-Effectiveness for Cantonese LLMs.}

High-quality Cantonese data is crucial for the pre-training or alignment of Cantonese LLMs, with translations from Standard Chinese proving more effective due to linguistic similarities (Figure~\ref{modeloverview2}i), as opposed to English (Table~\ref{TRANS_1},~\ref{TRANS_2}). While models like Gemma-2-27b perform less effectively in English-to-Cantonese translation, closed-source models such as Sensechat-5 and GPT series show minimal quality difference between 0-shot and 5-shot settings. Prioritizing translations from Standard Chinese, then English, optimizes data quality. Regarding cost-effectiveness, using closed-source models like Sensechat-5-Cantonese, ChatGPT, and GPT-4o is advisable if API costs are negligible (Table~\ref{TRANS_1},~\ref{TRANS_2}). Models like Mixtral-large-2, Llama-3.1-70b and Qwen-1.5-110b offer cost savings and high-quality translations in both settings (Table~\ref{Trans_time}, Figure~\ref{modeloverview2}i). The Llama and Qwen series, while not the highest in output quality, provides the best speed and cost-effectiveness for translating datasets to Cantonese.


\subsection{Case study}
In addition to the results analyzed above, we find that Gemma-2-27b frequently encounters hallucination issues, which impair its ability to handle Cantonese tasks (Appendix~\ref{case_study}). Although Qwen-2-72b exhibits good performance, it sometimes outputs training data. Nonetheless, the Qwen series of models remains proficient in handling Cantonese tasks (Appendix~\ref{case_study}). Appendix~\ref{case_study} for more cases.

\section{Challenges and opportunities}



\subsection{Existing Cantonese challenges}

\paragraph{Colloquialism.}

Cantonese differs significantly from Standard Chinese in its spoken vocabulary, posing unique challenges for NLP models initially trained on Mandarin~\cite{snow2004cantonese, xiang2024cantonese}. These differences are particularly evident in informal settings such as speech transcription and online forums like Linkg, and Openrice. Although smaller compared to datasets for English and Standard Chinese models like BERTweet~\cite{nguyen2020bertweet} and MacBERT~\cite{cui2021pre}, these platforms still provide a substantial text corpus for training Cantonese-specific models~\cite{hale2001probabilistic, hale2016information}.
The abundant unique expressions and slang in Cantonese, often embedded with complex cultural nuances, hinder adaptation of Standard Chinese-based models to Cantonese. For example, 
``Wan2 Sik6'' literally means ``looking for food'', but it is commonly used to describe seeking employment or earning money, carrying connotations of survival and making a living in Cantonese. 
In addition, common spelling mistakes and novel meanings in Cantonese further complicate model training, emphasizing the need for robust, Cantonese-specific vocabularies and corpora to capture the full breadth of colloquialisms and idioms of the language~\cite{li2009punning}.

\paragraph{Multilingualism.} To elucidate the multilingual dynamics in social media of Hong Kong, \cite{xiang2024cantonese} identify frequent code-switching between Cantonese and Standard Chinese, and a significant presence of English~\cite{yue1991yue, li2006maritime}. Highlighting the multilingualism, examples include Cantonese sentences incorporating English terms, such as ``deadline'' seamlessly integrated as in ``Gan2 M4 Cit3 deadline'' (struggling to meet the deadline), and the use of the Japanese loanword ``Kawaii'' (cute), pronounced and adapted locally in phrases like ``Ni1 Gin6 Saam1 Hou2 kawaii'' (This shirt is very cute). These findings underscore the need for Cantonese NLP systems to handle multilingual code-switching and suggest adding spelling correction and dialect identification to improve data processing.

\begin{figure}[h]
    \centering
    \includegraphics[width=8cm]{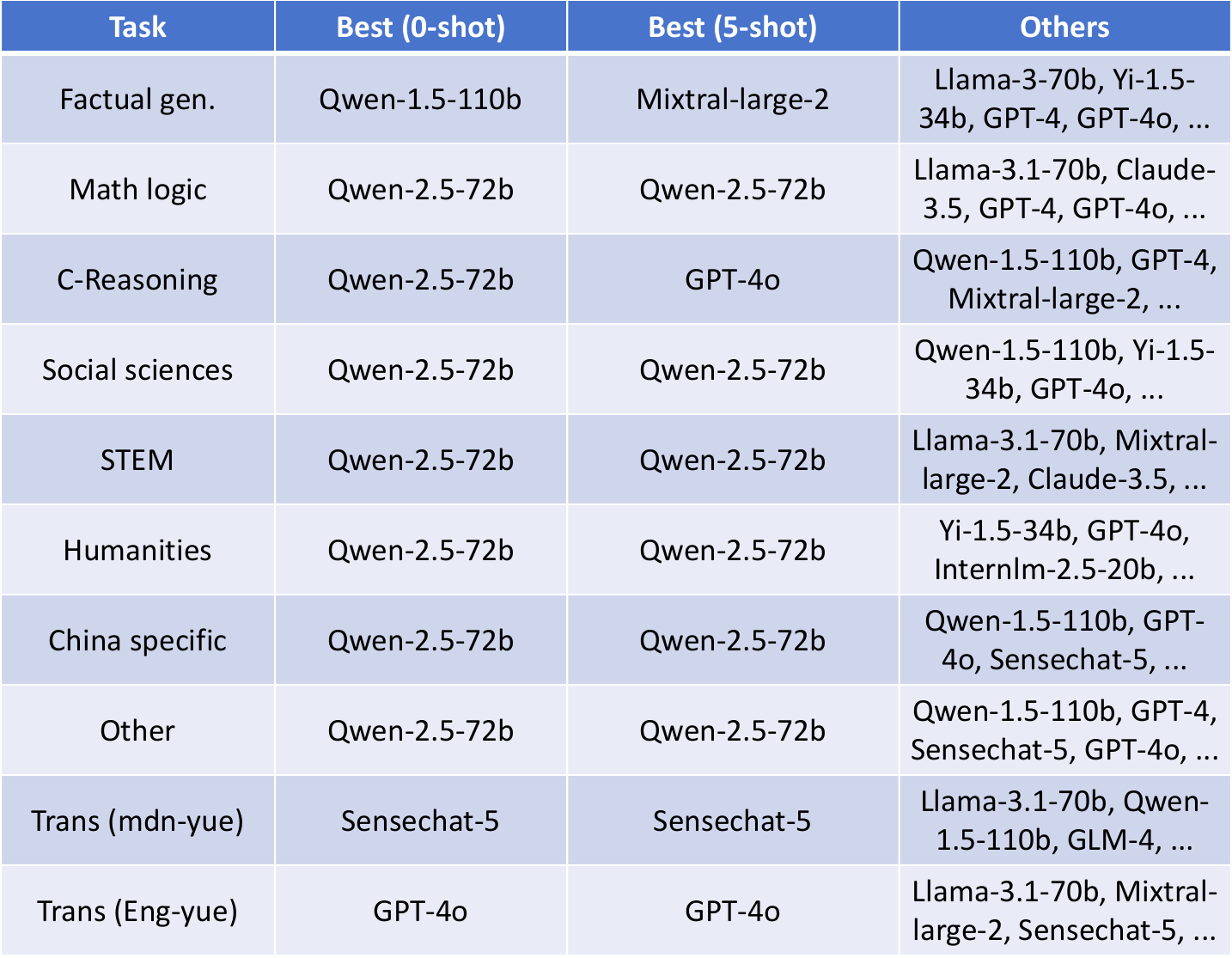}
    \caption{LLMs proficient in handling various tasks.}
    \label{Oppo}
\end{figure}

\subsection{Opportunities}
Given the existing challenges in Cantonese language and the evaluation results on benchmarks, we propose the following potential research directions and recommended models.

\textbf{Data augmentation.}
Data augmentation methods for Cantonese are similar to those used broadly, including label-invariant methods that modify text while preserving labels~\cite{weizou2019eda, min2020syntactic, shi2021substructure}, and label-variant techniques that alter semantics for new instances~\cite{jinetal2019imat, dai2019style}. Supervised contrastive learning enhances task-specific neural representations~\cite{sedghamizetal2021supclseq}, and LLM-based strategies are reviewed in~\cite{ding2024data}. For dataset conversion to Cantonese, high-capability models like Sensechat-5 and GPT-4 are recommended if costs allow (Table~\ref{TRANS_1},~\ref{TRANS_2}, Table~\ref{Trans_time}). Budget-friendly alternatives include Mixtral-large-2 and Llama-3.1-70b, with Llama models providing cost-effective speeds despite lower quality (Table~\ref{Trans_time}).


\textbf{Code-switch.}
Developments in LLMs suggest emergent abilities for untrained tasks, although effectiveness varies across scripts and languages~\cite{mann2020language, bang2023multitask}. Research in SCN-adapted LLMs is progressing, benefiting Cantonese NLP in the future~\cite{cui2023efficient, bai2023qwen}. We propose four benchmarks and compile a Yue-TRANS dataset, each involving two or more languages. Therefore, based on the performance observe on benchmarks, we recommend using newer versions of the Qwen, Llama, Mixtral, and Yi series (Figure~\ref{Oppo}). 

\textbf{Large language models.}
Based on the analysis above, we compile Figure~\ref{Oppo}, which presents the best LLMs in 0-shot and 5-shot, and suggests LLM series for various tasks. For work related to LLMs, we recommend using newer versions of the Qwen, Mixtral, Llama, and Yi series (Table~\ref{TruthfulQA_Cant_all},\ref{GSM8K_Cant_all},~\ref{ARC-C_Cant_all},~\ref{MMLU_Cant_all}). For tasks that involve only prompting, without the need for LLM training, we also recommend using closed-source models such as GPT, GLM, and Sensechat series models.

\section{Conclusion and Outlook}

Cantonese, spoken by over 85 million people, lags in natural language processing development, especially in large language models. To address this gap, we summarize existing Cantonese NLP methods and introduce four new benchmarks (Yue-Truthful, Yue-GSM8K, Yue-ARC-C, Yue-MMLU) and a translation dataset (Yue-TRANS). We evaluate 35 mainstream LLMs on these benchmarks, identifying current strengths and weaknesses. This work lays a foundation for advancing Cantonese LLM related technology.

Future efforts focus on building larger, high-quality Cantonese corpora and optimizing models for Cantonese-specific tasks. Collaboration among global researchers accelerates progress, helping Cantonese NLP catch up with other languages, enriching the experiences of Cantonese speakers.


\section*{Limitations}
The first limitation is the scarcity of work related to Cantonese LLMs, which restricts the extent of summarizing relevant studies. However, it is believed that with the publication of this paper, an increasing number of projects involving large-scale Cantonese models will be proposed. The second limitation is that the recommended LLMs presented in the article are for reference only; LLMs not recommended are not necessarily of inferior quality, nor does it imply they are unsuitable for Cantonese-related tasks. The selection of specific models for Cantonese-related tasks should be based on a detailed analysis of the specific issues at hand. 

In addition, we specifically focus on benchmarking vanilla LLMs without fine-tuning to test these LLMs' intrinsic abilities, which can also better inform their performance after fine-tuning. 

\section*{Ethics Statement}
Concerning the data annotators and the evaluation of data review, we ensure the selection of qualified tri-lingual individuals from Hong Kong and Guangdong who are compensated with reasonable hourly wages or other forms of subsidies as rewards. We have already obtained approval for this research from the Ethics Review Committee.


\section*{Acknowledgements}

We want to thank our anonymous AC and reviewers for their feedback. This work was supported by Hong Kong Innovation and Technology Commission’s Innovation and Technology Fund (Award No. ITS/269/22FP). 

\bibliography{custom}

\appendix

\section{Appendix}
\label{sec:appendix}

\subsection{Cantonese speaking population statistics}

\begin{table}[h]
    \centering
    \begin{tabular}{l|c|c}
        \toprule
        \textbf{Country/region} & \textbf{Pop.} & \textbf{Stat. Time} \\
        \midrule
        Hong Kong & 6,529,000 & 2021 \\
        Macau & 506,000 & 2016 \\
        Guangdong & 40,000,000 & 2021 \\
        Guangxi & 12,000,000 & 2022 \\
        Brunei & 6,350 & 2006 \\
        Indonesia & 180,000 & 1982 \\
        Malaysia & 1,070,000 & 2000 \\
        Philippines & 9,780 & 2000 \\
        Singapore & 338,000 & 1993 \\
        Thailand & 29,400 & 1984 \\
        Vietnam & 862,000 & 1999 \\
        \bottomrule
    \end{tabular}
    \caption{Cantonese speaking population statistics. Pop. is population. Stat. Time is statistical time}
    \label{stat_all}
\end{table}

\subsection{Existing Cantonese data}

At the end of the 16th century, Matteo Ricci compiles the first ``Modern Bilingual Chinese Dictionary'', significantly incorporating Cantonese terms, highlighting its role in Sino-Western interactions. By the 19th century, most bilingual dictionaries focus on Cantonese \cite{xiang2024cantonese}. Historically, Hong Kong and related institutions lead Cantonese data initiatives. \cite{wu1994aligning} creates a bilingual parallel corpus from the Hong Kong Legislative Council records, in both Standard Chinese and English. This effort is complemented by \cite{hun1999cancorp}, who pioneers a Cantonese-only corpus with one million characters from dialogues involving children in Hong Kong, and \cite{yip2007bilingual}, who develops a bilingual corpus for Cantonese-speaking children. Additionally, a notable Cantonese corpus comes from television and theatrical productions in Hong Kong \cite{leung2001hkcac}. The University of Hong Kong further contributes by collecting and annotating spontaneous speech from dialogues and broadcasts, focusing on segmentation, parts of speech tagging, and phonetic transcription \cite{ping2006specification}. \cite{lee2011toward} introduces a parallel corpus for machine translation between Cantonese and Standard Chinese, aligned at the sentence level, using data from Cantonese speeches on Hong Kong television and their Standard Chinese subtitles.

Recent efforts aim to bridge the data gap between Cantonese and other major languages. These include a small parallel dependency treebank for Cantonese and Mandarin, with 569 aligned sentences annotated using the Universal Dependencies scheme, and excerpts from the ``ABC Cantonese-English Comprehensive Dictionary'' providing 14,474 high-quality Cantonese-English parallel sentences, crucial for translation system development.

\subsection{Cantonese small-scale neural network}

Cantonese NLP research spreads across various topics, including rumor detection, sentiment analysis, machine translation, dialogue. We collect existing small neural network methods, models, and tools.

\paragraph{Rumor detection.}
\cite{Chen2020} develop a dataset of 27,328 Cantonese tweets for rumor detection, split into 13,883 rumors and 13,445 non-rumors. They introduce an attention-based model, XGA, which combines XLNet~\cite{yang2019xlnet} and BiGRU to analyze both semantic and sentiment aspects. \cite{chen2024deep} develop CantoneseBERT to capture glyph and pronunciation clues of Cantonese characters, and introduces a Cantonese rumor detection model, SA-GCN, that encodes global structural information of tweet hierarchies using the BiGCN model and extracts local semantic features with the CantoneseBERT model. 

\paragraph{Sentiment analysis.}
Cantonese sentiment analysis utilizes diverse methodologies to address its linguistic complexities. \cite{zhang2011sentiment} apply Naive Bayes and SVMs with character-based bi-grams in the Openrice app for effective emotion detection. \cite{chen2013sentiment, chen2015opinion} deploy Hidden Markov Models for text segmentation and part-of-speech tagging, developing emotion-specific dictionaries via rule-based systems. These studies demonstrate the value of combining machine learning with lexical techniques~\cite{zhang2011sentiment, chen2013sentiment, chen2015opinion}. In addition, \cite{ngai2018multiple} and \cite{xiang2019sentiment} enhance classification accuracy using supervised and unsupervised methods in various domains. \cite{lee2019emotion} explores fine-grained emotion analysis across languages, achieving significant results. These efforts underscore the importance of multi-methodological approaches~\cite{ngai2018multiple, xiang2019sentiment, lee2019emotion}. \cite{tan2021analysis} successfully employ Transformers pre-trained on simplified Chinese~\cite{tan2021analysis}.

\paragraph{Machine translation.}
Initial research in this area utilizes heuristic rules, with significant contributions from \cite{zhang1998dialect} and a bilingual Cantonese-English knowledge base by \cite{wu2006structural}. The focus has since shifted to statistical machine translation, exemplified by \cite{huang2016machine}, who addresses the challenges of translating between Cantonese and Mandarin with limited resources. \cite{wong2018register} improves this approach by enhancing parallel data for more efficient model training. Recent developments include a large-scale evaluation dataset by \cite{liu2022low}, containing over 35,000 Mandarin-Cantonese sentence pairs, and unsupervised translation models by \cite{dare2023unsupervised}, which use cross-lingual embeddings and combine Transformer architecture with character-based tokenization to create a new corpus of approximately 1 million Cantonese sentences.

\paragraph{Dialogue summarization and generation.}
\cite{lee2021restatement} explores generating questions and restating information in Cantonese dialogue systems, particularly for counseling chatbots. They enhance performance by fine-tuning the pre-trained BertSum model \cite{liu2019text} on Cantonese data, effective in tasks involving text summarization and question generation. In dialogue generation, \cite{lee2021response} develops a specialized dataset for virtual counselors containing 1,028 post-reply pairs addressing test anxiety and loneliness, using these categories to guide response selection through a regression model.

\paragraph{Cantonese language model.}
Training Cantonese language models like XLNet \cite{yang2019xlnet} and ELECTRA \cite{clark2020electra} from ToastyNews\footnote{\url{https://huggingface.co/toastynews}} faces challenges due to data scarcity and legal constraints. \cite{chen2024deep} introduce CantoneseBERT and the SA-GCN model for detailed analysis and rumor detection in tweets, utilizing innovative methods like permutation learning and adversarial training. However, the training corpus largely includes Standard Chinese, leading to potential language contamination, and the impact on model efficacy remains unexplored.

\paragraph{Cantonese NLP tools.}

The landscape of Cantonese NLP tools is diverse, addressing various needs. PyCantonese~\cite{lee2022pycantonese} facilitates corpus data handling and linguistic analysis. Hong Kong Cantonese Localization provides culturally contextual translations. TransCan\footnote{\url{https://github.com/ayaka14732/TransCan}} enhances English-to-Cantonese translation, surpassing commercial solutions like Baidu and Bing. Text segmentation tools like Cantonese Word Segmentation\footnote{\url{https://github.com/wchan757/Cantonese_Word_Segmentation}} and cantoseg\footnote{\url{https://github.com/ayaka14732/cantoseg}} improve accuracy through custom dictionaries. canto-filter\footnote{\url{https://github.com/CanCLID/canto-filter}} categorizes texts based on linguistic features, while songotsti\footnote{\url{https://github.com/justinchuntingho/songotsti}} and fast-langid\footnote{\url{https://github.com/ffreemt/fast-langid}} offer additional support for language identification.

\subsection{Cantonese large language model}

Developing Cantonese LLMs is challenging due to scarce resources and the distinct features of the Cantonese language, necessitating extensive high-quality datasets for pre-training\footnote{\url{https://www.sensetime.com/en/news-detail/51168164?categoryId=1072}}. Despite these obstacles, these models show promising capabilities in processing Cantonese.

Aligning Cantonese LLMs for downstream tasks, such as prompting, supervised fine-tuning, and reinforcement learning from human feedback, is cost-effective and helps eliminate biases and meet cultural expectations.

Recent studies~\cite{fu2024efficacy} validate ChatGPT's effectiveness in Cantonese dialogue and sentiment analysis, analyzing messages from a Hong Kong web counseling service. The CanChat bot, introduced to enhance counseling services in Hong Kong, provides initial support to students, improving their emotional well-being during and beyond the COVID-19 pandemic~\cite{10361241}.

Regarding the training and reasoning technologies for LLMs associated with mainstream languages, there is no development specific to Cantonese such as LoRA~\cite{hu2021lora,wang2024lora, wang2024mos}, reasoning~\cite{gao2024interpretable, havrilla2024glore}, etc.

Transitioning from small-scale networks to exploring Cantonese LLMs, both general-purpose and closed-source models show promise, but quantifying performance is challenging. We propose four benchmarks to evaluate and advance Cantonese capabilities in LLMs.




\subsection{Evaluation tools}
\label{ealtools}
\begin{itemize}
\item \textbf{Rouge-l:} from rouge\_metric import PyRouge
\item \textbf{Bleu-4:} from nltk.translate.bleu\_score import sentence\_bleu, SmoothingFunction
\item \textbf{BERTScore:} bert-base-multilingual-cased \& roberta-large
\end{itemize}


\subsection{Yue-MMLU}
\label{yue-mmlu}
We select twenty-two topics from CMMLU that cover most of the themes in CMMLU to serve as the topics for Yue-MMLU, which are as follows: 
\begin{itemize}
\item chinese\_civil\_service\_exam
\item arts
\item electrical\_engineering
\item chinese\_literature
\item education
\item economics
\item ethnology
\item college\_medicine
\item journalism
\item management
\item marketing
\item philosophy
\item security\_study
\item sociology
\item world\_history
\item world\_religions
\item high\_school\_geography
\item machine\_learning
\item marxist\_theory
\item professional\_psychology
\item sports\_science
\item logical
\end{itemize}

\subsection{Source of evaluation LLMs}
\label{source_llm}
This section covers the evaluation of LLMs along with the corresponding Hugging Face links and the names of the APIs.

\subsection{Experimental results}

\subsubsection{Cantonese and English TruthfulQA (best and incorrect)}
Table~\ref{TruthfulQA_Can_best} (comparison between \textbf{best} answer and groundtruth) and Table~\ref{TruthfulQA_incorrect} (comparison between \textbf{incorrect} answer and groundtruth) are the experimental results based on the Cantonese and English version of TruthfulQA.

\subsubsection{English TruthfulQA (correct)}
Table~\ref{TruthfulQA_Eng} (comparision vetween \textbf{correct} answer and groundtruth) is the experimental result based on the English version of TruthfulQA, intended for comparison with the Cantonese version of TruthfulQA. For more results, please refer to the publicly available evaluation platform\footnote{\url{https://huggingface.co/open-llm-leaderboard}}.

\subsubsection{English GSM8K}
Table~\ref{GSM8K_Eng} is the experimental result based on the English version of GSM8K, intended for comparison with the Cantonese version of GSM8K. For more results, please refer to the publicly available evaluation platform\footnote{\url{https://huggingface.co/open-llm-leaderboard}}.

\subsubsection{English ARC challenge}
Table~\ref{ARC_Eng} is the experimental result based on the English version of ARC Challenge, intended for comparison with the Cantonese version of ARC Challenge. For more results, please refer to the publicly available evaluation platform\footnote{\url{https://huggingface.co/open-llm-leaderboard}}.

\subsubsection{CMMLU}
Table~\ref{CMMLU} is the experimental result based on the Standard Chinese version of MMLU, intended for comparison with the Cantonese version of MMLU. For more results, please refer to the publicly available evaluation platform\footnote{\url{https://huggingface.co/open-llm-leaderboard}}.


\subsubsection{Translation}
Table~\ref{TRANS_1},~\ref{TRANS_2} is the experimental result based on the Yue-Trans datasets. Table~\ref{Trans_time_all} and Table~\ref{Trans_time} reflect the running time of different LLMs on the translation dataset.

\subsection{Prompt templates for multilingual evaluation}
\label{prompts}
This section details the prompt templates used for the Cantonese, English, and Standard Chinese datasets tested in our experiments. Each dataset was evaluated under both 0-shot and 5-shot settings. For the 5-shot setting, except for the translation task (Yue-TRANS), the prompts were generated using a sliding window approach, where the preceding five examples from the dataset (Yue-TruthfulQA, Yue-GSM8K, Yue-ARC-Challenge, and Yue-MMLU) were utilized as context for each new example. For the Yue-TRANS translation task, the BM25 algorithm was employed to identify and select the five most similar examples to serve as few-shot examples. Below, we outline the prompt structures and methodologies used for generating the few-shot examples.

The inference and evaluation processes in this study were facilitated by the OpenCompass platform, which provided a robust and universal evaluation framework for foundation models \cite{2023opencompass}.

\begin{CJK}{UTF8}{bsmi}
\subsubsection{Yue-TruthfulQA prompt}
\textbf{0-shot:}
\begin{Verbatim}
    用粵語答下面問題：
    問題：[QUESTION]
    回應：
\end{Verbatim}

\textbf{\hspace*{-0.4cm}5-shot:}
\begin{Verbatim}
    樣例1-5：
        問題：[EXAMPLE_QUESTION]
        回應：[EXAMPLE_ANSWER]
    
    用粵語答下面問題：
    問題：[TARGET_QUESTION]
    回應：
\end{Verbatim}
\end{CJK}

\subsubsection{En-TruthfulQA prompt}
\textbf{0-shot:}
\begin{Verbatim}
    Answer the following question in English:
    Question: [QUESTION]
    Answer:
\end{Verbatim}

\textbf{\hspace*{-0.4cm}5-shot:}
\begin{Verbatim}
    Example 1-5:
        Question: [EXAMPLE_QUESTION]
        Answer: [EXAMPLE_ANSWER]
    
    Answer the following question in English:
    Question: [TARGET_QUESTION]
    Answer: 
\end{Verbatim}

\begin{CJK}{UTF8}{bsmi}
\subsubsection{Yue-GSM8K prompt}

\textbf{0-shot:}
\begin{Verbatim}
    請逐步思考，最終答案前用「####」標記。用粵語答下面問題：
    問題：[QUESTION]
    用粵語回答問題：
\end{Verbatim}

\textbf{\hspace*{-0.4cm}5-shot:}
\begin{Verbatim}
    樣例1-5：
        問題：[EXAMPLE_QUESTION]
        回應：[EXAMPLE_ANSWER]
    
    請逐步思考，最終答案前用「####」標記。用粵語答下面問題：
    問題：[TARGET_QUESTION]
    用粵語回答問題：
\end{Verbatim}
\end{CJK}

\subsubsection{En-GSM8K prompt}

\textbf{0-shot:}
\begin{Verbatim}
    Please think step by step, mark the final answer with '####'.
    Answer the following question in English:
    Question: [QUESTION]
    Answer the question in English:
\end{Verbatim}

\textbf{\hspace*{-0.4cm}5-shot:}
\begin{Verbatim}
    Example 1-5:
        Question: [EXAMPLE_QUESTION]
        Response: [EXAMPLE_ANSWER]
    
    Please think step by step, mark the final answer with '####'.
    Answer the following question in English:
    Question: [TARGET_QUESTION]
    Answer the question in English:
\end{Verbatim}

\begin{CJK}{UTF8}{bsmi}
\subsubsection{Yue-ARC-C prompt}

\textbf{0-shot:}
\begin{Verbatim}
    問題：[QUESTION]
    由提供嘅選項中直接用選項嘅字母作答。
    回應：
\end{Verbatim}

\textbf{\hspace*{-0.4cm}5-shot:}
\begin{Verbatim}
    樣例1-5：
        問題：[EXAMPLE_QUESTION]
        回應：[EXAMPLE_ANSWER]
    
    問題：[TARGET_QUESTION]
    由提供嘅選項中直接用選項嘅字母作答。
    回應：
\end{Verbatim}
\end{CJK}

\subsubsection{En-ARC-C prompt}

\textbf{0-shot:}
\begin{Verbatim}
    Question: [QUESTION]
    Answer with the option's letter from the given choices directly.
    Answer:
\end{Verbatim}

\textbf{\hspace*{-0.4cm}5-shot:}
\begin{Verbatim}
    Example 1-5:
        Question: [EXAMPLE_QUESTION]
        Answer: [EXAMPLE_ANSWER]
    
    Question: [TARGET_QUESTION]
    Answer with the option's letter from the given choices directly.
    Answer:
\end{Verbatim}

\begin{CJK}{UTF8}{bsmi}
\subsubsection{Yue-MMLU prompt}

\textbf{0-shot:}
\begin{Verbatim}
    以下係關於[SUBJECT]嘅單項選擇題，請直接畀出正確答案嘅選項。
    問題：[QUESTION]
    答案：
\end{Verbatim}

\textbf{\hspace*{-0.4cm}5-shot:}
\begin{Verbatim}
    樣例1-5：
        問題：[EXAMPLE_QUESTION]
        回應：[EXAMPLE_ANSWER]
    
    以下係關於[SUBJECT]嘅單項選擇題，請直接畀出正確答案嘅選項。
    問題：[TARGET_QUESTION]
    答案：
\end{Verbatim}
\end{CJK}

\begin{CJK}{UTF8}{gbsn}
\subsubsection{Zh-CMMLU prompt}

\textbf{0-shot:}
\begin{Verbatim}
    以下是关于[SUBJECT]的单项选择题，请直接给出正确答案的选项。
    题目：[QUESTION]
    答案：
\end{Verbatim}

\textbf{\hspace*{-0.4cm}5-shot:}
\begin{Verbatim}
    样例1-5:
        题目：[EXAMPLE_QUESTION]
        答案：[EXAMPLE_ANSWER]
    
    以下是关于[SUBJECT]的单项选择题，请直接给出正确答案的选项。
    题目：[TARGET_QUESTION]
    答案：
\end{Verbatim}
\end{CJK}

\begin{CJK}{UTF8}{bsmi}
\subsubsection{Yue-TRANS prompt}
\textbf{0-shot:}
\begin{Verbatim}
    請將下面呢句/段話直接翻譯成粵語：[SOURCE_TEXT]
\end{Verbatim}

\textbf{\hspace*{-0.4cm}5-shot:}
\begin{Verbatim}
    樣例1-5：
        請將下面呢句/段話直接翻譯成粵語：[EXAMPLE_SOURCE_TEXT]
        翻譯：[EXAMPLE_TRANSLATION_TEXT]
    
    根據上面嘅例子，請將下面呢句/段話直接翻譯成粵語：
    [TARGET_SOURCE_TEXT]
\end{Verbatim}
\end{CJK}

\section{Result}

\begin{table*}\small
\centering
\begin{tabular}{l|c|c|c|c|c|c}
\toprule
\multirow{2}{2.7cm}{\textbf{Models \\ (Yue-TruthfulQA)}} & \multicolumn{3}{c|}{\textbf{0-shot (correct)}} & \multicolumn{3}{c}{\textbf{5-shot (correct)}} \\
\cmidrule{2-7}
& \textbf{Rouge-l} & \textbf{Bleu-4} & \textbf{BERTScore}  & \textbf{Rouge-l} & \textbf{Bleu-4} & \textbf{BERTScore}  \\
\midrule
Qwen-7b & 6.42  & 3.99 & 51.57 & 4.04 	&	2.98  & 49.7 \\
Qwen-1.5-7b & 20.54 	& 	13.41  & 66.45 & 12.45 	& 	10.41  & 61.59 \\
Qwen-1.5-110b & 26.04 & 15.95 & 69.29 & 31.73 & 19.53 & 70.87 \\
Qwen-2-7b & 13.27 & 10.00 & 66.14  & 16.91 & 11.48 & 67.71  \\
Qwen-2-72b & 10.86 & 9.68 & 65.62  & 17.52 & 12.38 & 67.72  \\
Qwen-2.5-7b & 18.51 	& 	12.28  & 66.07  & 6.83 	& 	8.07  & 58.97\\
Qwen-2.5-72b & 13.03 	& 	9.64   & 66.94 & 20.23 	& 	12.87   & 69.53 \\
Mixtral-8x22b & 14.74 & 10.83 & 66.72  & 20.40 & 14.09 & 68.05 \\
Mixtral-large-2 & 19.72 & 13.01 & 69.06 & 31.38 & 18.61 & 72.07 \\
Llama-2-7b & 3.48 	& 	6.42  & 57.16 & 3.57 	& 	6.52  & 56.36 \\
Llama-3-8b & 8.40 & 8.68 & 64.37 & 28.68 & 16.43 & 70.82 \\
Llama-3-70b & 10.98 & 9.51 & 66.10 & 33.06 & 19.31 & 71.95 \\
Llama-3.1-8b & 13.82 & 10.33 & 66.97 & 26.18 & 15.20 & 70.28 \\
Llama-3.1-70b & 21.03 & 14.30 & 68.31 & 34.72 & 20.54 & 70.80 \\
Phi-3-medium & 18.70 & 12.00 & 67.36 & 22.00 & 13.72 & 67.57 \\
Gemma-2-27b & 8.09 & 8.44 & 64.41 & 11.33 & 9.98 & 63.66  \\
Yi-6b & 1.37 	& 	5.05  & 53.16  & 1.07 	& 	5.99  & 54.21 \\
Yi-1.5-6b & 1.21 	& 	4.60   & 42.15 & 1.04 	& 	6.15  & 53.85  \\
Yi-1.5-34b & 15.41 & 11.11 & 67.57 & 20.30 & 13.20 & 69.50 \\
Internlm-7b & 5.89 & 6.65 & 56.33 & 2.59  & 3.68 & 55.73  \\
Internlm-7b-turbomind & 5.91 & 6.71 & 56.71 & 2.77  & 3.82 & 55.57 \\
Internlm-2-7b & 7.93 & 10.21  & 63.81 & 17.66 & 16.62 & 33.33  \\
Internlm-2-7b-chat & 6.7 & 7.68 & 61.83 & 3.3 & 5.49 & 65.47  \\
Internlm-2-7b-turbomind & 8.09   & 10.53 & 64.3 & 17.69 & 16.99 & 63.68  \\
Internlm-2.5-7b & 8.96 & 10.53 & 66.11 & 10.3 & 14.47 & 67.73 \\
Internlm-2.5-7b-chat & 7.13 & 8 & 63.48 & 4.05 & 7.19 & 67.61 \\
Internlm-2.5-7b-turbomind & 8.93 & 10.46 & 65.75 & 10.12 & 14.39 & 67.14 \\
Internlm-2.5-20b-chat & 6.96 & 7.73 & 62.99 & 3.28 & 6.06   & 66.99 \\
Internlm-2.5-20b-turbomind & 9.49  & 11.55  & 66.70 & 11.98  & 16.56   & 68.86\\
ERNIE-Lite & 20.58 & 12.23 & 67.64 & 20.69 & 12.27 & 68.45 \\
ERNIE-Tiny & 27.16 & 14.49 & 68.45 & 27.91 & 15.28 & 68.84 \\
ERNIE-Speed & 22.58 & 13.15 & 67.84 & 23.61 & 13.82 & 68.27 \\
ERNIE-Turbo & 17.91 & 11.30 & 66.71  & 21.19 & 12.19 & 68.29 \\
Sensechat-5 & 24.75 & 15.11 & 68.43 & 32.45 & 19.70 & 70.02  \\
Claude-3.5 & 14.23 & 9.95 & 67.56 & 12.66 & 10.06 & 68.12  \\
GLM-4 & 13.44 & 10.07 & 67.26 & 23.57 & 14.28 & 70.30 \\
ChatGPT & 25.07 & 14.81 & 67.78 & 31.84 & 18.42 & 70.41 \\
GPT-4o & 17.58 & 12.17 & 68.68 & 27.64 & 16.52 & 71.59 \\
GPT-4 & 19.47 & 13.45 & 68.99 & 28.43 & 16.74 & 71.26 \\
\bottomrule
\end{tabular}
\caption{Results of the comparison between texts generated by various LLMs in Yue-TruthfulQA based on 0-shot and 5-shot settings and the correct texts. \textbf{Rouge-l}, \textbf{Bleu-4}, and \textbf{BERTScore} are evaluation metrics for comparing text similarity.}
\label{TruthfulQA_Cant_all}
\end{table*}

\begin{table*}[h]\small
\centering
\begin{tabular}{l|c|c}
\toprule
\textbf{Models} & \textbf{Acc. (0-shot)} & \textbf{Acc. (5-shot)} \\
\midrule
Qwen-7b & 0.68    & 6.75  \\
Qwen-1.5-7b & 36.62 	&26.31   \\
Qwen-1.5-110b & 54.89 &	58.30  \\
Qwen-2-7b & 50.49 &	61.11  \\
Qwen-2-72b & 77.86 &	77.71  \\
Qwen-2.5-7b & 63.84 &	44.20   \\
Qwen-2.5-72b & 83.62 &	83.55   \\
Mixtral-8x22b & 65.20 &	66.19  \\
Mixtral-large-2 & 80.14 &	81.27  \\
Llama-2-7b & 0.83 	&1.82  \\
Llama-3-8b & 52.46 &	49.66  \\
Llama-3-70b & 73.62 &	75.66  \\
Llama-3.1-8b & 63.91 &	61.64  \\
Llama-3.1-70b & 53.60 &	79.00  \\
Phi-3-medium & 59.29 &	63.15  \\
Gemma-2-27b & 9.70 &	2.65  \\
Yi-6b & 2.12 &	10.16  \\
Yi-1.5-6b & 3.94 &	3.49  \\
Yi-1.5-34b & 69.45 &	69.45  \\
Internlm-7b-turbomind & 4.55 &	9.48  \\
Internlm-2-7b & 11.90 &	22.21  \\
Internlm-2-7b-chat & 56.41 &	48.67  \\
Internlm-2-7b-turbomind & 11.37 	&23.96  \\
Internlm-2-20b & 12.81 &	8.87  \\
Internlm-2-20b-chat & 60.42 &	59.21  \\
Internlm-2.5-7b & 57.70 &	44.05  \\
Internlm-2.5-7b-chat & 65.96 &	64.67  \\
Internlm-2.5-7b-turbomind & 56.79 &	42.99  \\
Internlm-2.5-20b-chat & 71.87 &	72.33   \\
Internlm-2.5-20b-turbomind & 45.03 &	61.41  \\
ERNIE-turbo & 14.03 	&10.92  \\
ERNIE-Speed & 28.81 	&28.28  \\
ERNIE-Lite & 54.81 	&32.15  \\
ERNIE-Tiny & 2.73 &	3.94  \\
\textbf{SenseChat-5} & 77.48 & 73.16 \\
Claude-3.5 & 77.79 &	81.27  \\
GLM-4 & 78.17 &	77.10  \\
ChatGPT & 23.35&	41.09 \\
GPT-4o & 83.24 &	83.40  \\
GPT-4 & 81.12 &	83.02  \\
\bottomrule
\end{tabular}
\caption{Results of the comparison between answer generated by various LLMs in Yue-GSM8K based on 0-shot and 5-shot settings and groundtruth.}
\label{GSM8K_Cant_all}
\end{table*}

\begin{table*}[h]\small
\centering
\begin{tabular}{l|c|c}
\toprule
\textbf{Models} & \textbf{Acc. (0-shot)} & \textbf{Acc. (5-shot)} \\
\midrule
Qwen-7b & 11.02	&14.6  \\
Qwen-1.5-7b & 65.24	&67.55  \\
Qwen-1.5-110b & 88.64	&90.09 \\
Qwen-2-7b & 79.08&	78.39  \\
Qwen-2-72b & 88.64	&88.56  \\
Qwen-2.5-7b & 81.64&	83.35  \\
Qwen-2.5-72b & 92.74&	92.91  \\
Mixtral-8x22b & 76.09&	76.09 \\
Mixtral-large-2 & 89.5&	90.61 \\
Llama-2-7b & 23.57	&34.24  \\
Llama-3-8b & 70.11&	53.8 \\
Llama-3-70b & 85.06&	84.97 \\
Llama-3.1-8b & 69	&67.81 \\
Llama-3.1-70b & 88.98	&88.39 \\
Phi-3-medium & 77.63&	78.31 \\
Gemma-2-27b & 67.98&	55.59 \\
Yi-6b & 31&	66.01 \\
Yi-1.5-6b & 34.59&	66.7 \\
Yi-1.5-34b & 84.88	&86.42 \\
Internlm-7b-turbomind & 44.75&	55.34 \\
Internlm-2-7b-turbomind & 44.75	&55.34 \\
Internlm-2.5-7b & 78.14	&77.46 \\
Internlm-2.5-7b-chat & 81.21	&79.85 \\
Internlm-2.5-7b-turbomind & 77.37&	77.37 \\
Internlm-2.5-20b-chat & 82.15	&82.58 \\
Internlm-2.5-20b-turbomind & 84.29&	76.94 \\
ERNIE-turbo & 44.41	&46.46 \\
ERNIE-Speed & 74.47	&74.04 \\
ERNIE-Lite & 72.25	&77.28 \\
ERNIE-Tiny & 34.67&	32.88 \\
SenseChat-5 & 88.47&	87.28 \\
Claude-3.5 & 91.55&	92.23 \\
GLM-4 & 88.9&	88.73 \\
ChatGPT & 69.68&	70.71 \\
GPT-4o & 91.97&	94.45 \\
GPT-4 & 92.66&	92.06 \\
\bottomrule
\end{tabular}
\caption{Results of the comparison between answer generated by various LLMs in Yue-ARC-C based on 0-shot and 5-shot settings and groundtruth.}
\label{ARC-C_Cant_all}
\end{table*}

\begin{table*}\small
\centering
\begin{tabular}{l|c|c|c|c|c|c|c|c|c|c}
\toprule
\multirow{2}{2cm}{\textbf{Models \\ (Yue-MMLU)}} & \multicolumn{5}{c|}{\textbf{0-shot (correct)}} & \multicolumn{5}{c}{\textbf{5-shot (correct)}} \\
\cmidrule{2-11}
& \textbf{STEM} & \textbf{Hum.} & \textbf{S.S.} & \textbf{C.S.} & \textbf{Oth.} & \textbf{STEM} & \textbf{Hum.} & \textbf{S.S.} & \textbf{C.S.} & \textbf{Oth.}\\
\midrule
        
        Qwen-7b & 10.1& 12.95& 12.12& 11.61& 7.96& 9.98& 15.96& 14.48& 13.33& 13.26 \\ 
        Qwen-1.5-7b & 46.28& 61.65& 56.57& 50.02& 53& 60.14& 70.09& 65.55& 58.31& 65.02 \\ 
        Qwen-1.5-110b & 75.07& 88.48& 83.89& 80.57& 82.14& 79.96& 88.12& 88.75& 84.8& 89.31 \\ 
        Qwen-2-7b & 70.06& 81.04& 80.07& 69.54& 76.04& 74.08& 80.45& 80.7& 73.7& 79.52 \\ 
        Qwen-2-72b & 81.68&89.93&88.47&81.9&87.48&85.7&89.54&88.12&83.72&87.73 \\ 
        Qwen-2.5-7b & 72.86	&81.66&	78.25&	66.56&	75.19&	78.05&	80.37&	78.99&	69.82&	78.86 \\ 
        Qwen-2.5-72b & 83.72&	87.88&	87.2&	80.68&	85.36&	83.89&	89.7&	88.75&	82.34&	87.42 \\ 
 Mixtral-8x22b & 50.4& 57.08& 59.28& 44.02& 48.76& 58.94& 59.72& 62.44& 49.78& 57.83 \\ 
        Mixtral-large-2 & 60.38& 76.08& 74.92& 60.19& 70.74& 68.5& 79.65& 78.84& 63.85& 71.66 \\ 
        Llama-2-7b & 23.34& 23.84& 23.76& 22.78& 24.52& 27.48& 30.4& 31.76& 28.9& 24.38 \\ 
        Llama-3-8b & 49.13& 59.3& 56.51& 47.53& 53.72& 44.04& 58.47& 53.94& 46.24& 52.55 \\ 
        Llama-3-70b & 65.17& 73.58& 75.22& 57.87& 72.84& 64.06& 72.82& 73.16& 57.34& 72.95 \\ 
        Llama-3.1-8b & 45.96& 58.27& 56.08& 44.86& 53.7& 53.45& 58.06& 58.31& 45.86& 53.65 \\ 
        Llama-3.1-70b & 67.32& 76.57& 76.93& 60.96& 73.56& 72.23& 78.13& 78.23& 64.16& 74.9 \\ 
        Phi-3-medium & 45.26& 61.42& 58.4& 45.65& 51.33& 49.88& 59.33& 59.35& 45.49& 53.02 \\ 
        Gemma-2-27b & 48.5& 54.05& 53.32& 36.92& 48.22& 40.62& 41.72& 43.81& 32.99& 46.03 \\ 
        Yi-6b & 36.46& 67.62& 57.32& 57.42& 50.06& 58.11& 72.14& 68.4& 60.56& 68.46 \\ 
        Yi-1.5-6b & 17.34& 35.98& 38.77& 32.9& 25& 58.53& 67.89& 66.56& 60& 62.05 \\ 
        Yi-1.5-34b & 68.48& 81.92& 81.74& 70.89& 79.76& 74.13& 85.12& 83.38& 78.2& 80.3 \\ 
        Internlm-7b-turbomind & 31.9& 48.79& 44.03& 41.14& 39.82& 39.84& 51.74& 50.06& 43.6& 42.32 \\ 
        Internlm-2-7b-turbomind & 51.69& 70.92& 64.71& 59.31& 58.93& 53.11& 68.51& 62.68& 59.77& 58.14 \\ 
        Internlm-2.5-7b & 65.34& 82.43& 79.24& 73.11& 74.15& 66.73& 81.06& 77.8& 71.65& 75.37 \\ 
        Internlm-2.5-7b-chat & 64.4& 80.92& 76.8& 70.24& 75.02& 65.04& 80.84& 76.79& 70.47& 75.19 \\ 
        Internlm-2.5-7b-turbomind & 65.34& 82.43& 79.24& 73.11& 74.15& 66.73& 81.06& 77.8& 71.65& 75.37 \\ 
        Internlm-2.5-20b-chat & 67.16& 81.56& 77.72& 73.05& 72.64& 66.22& 82.65& 78.42& 72.94& 74.03 \\ 
        Internlm-2.5-20b-turbomind & 72.86& 86.1& 82.14& 79.06& 74.7& 69.65& 78.79& 76.56& 70.28& 77.2 \\ 
        ERNIE-Lite & 53.45& 67.56& 67.73& 61.21& 61.21& 60.74& 70.27& 71.5& 62.43& 64.84 \\ 
        ERNIE-Tiny & 34.78& 37.86& 37.88& 33.08& 32.29& 32.52& 38.63& 37.58& 32.52& 34.6 \\ 
        ERNIE-turbo & 43.34& 56.05& 53.97& 52.02& 44.82& 41.01& 57.66& 54.28& 49.49& 46.95 \\ 
        Sensechat-5 & 69.97& 83.21& 80.73& 73.86& 76.95& 68.98& 82& 79.88& 73.52& 74.77 \\ 
        Claude-3.5 & 66.47& 76.84& 78.04& 60.6& 75.98& 75.92& 81.65& 84.24& 62.83& 82.54 \\ 
GLM-4 & 64.23& 84.39& 80.06& 75.66& 75.75& 72.18& 84.2& 80.07& 76& 78.06 \\ 
        ChatGPT & 49.78& 58.13& 58.74& 45.46& 52.42& 60.28& 59.81& 60.61& 47.5& 54.54 \\ 
        GPT-4o & 74.16& 83.28& 84.12& 71.6& 84.32& 72.35& 85.03& 84.32& 72.74& 81.58 \\ 
        GPT-4 & 67.68& 75.29& 77.26& 60.12& 74.46& 71.19& 76.75& 77.56& 63.5& 74.57 \\
        
\bottomrule
\end{tabular}
\caption{Results of the comparison between texts generated by various LLMs in Yue-MMLU based on 0-shot and 5-shot settings and the correct texts.}
\label{MMLU_Cant_all}
\end{table*}

\begin{table*}[h]\small
\centering
\begin{tabular}{l|c|c}
\toprule
\textbf{Models} & \textbf{Mode} & \textbf{Huggingface link \& API name} \\
\midrule
Qwen-7b & Huggingface & \url{https://huggingface.co/Qwen/Qwen-7B}  \\
\midrule
Qwen-1.5-7b & Huggingface & \url{https://huggingface.co/Qwen/Qwen1.5-7B}  \\
\midrule
Qwen-1.5-110b & Huggingface & \url{https://huggingface.co/Qwen/Qwen1.5-110B} \\
\midrule
Qwen-2-7b & Huggingface & \url{https://huggingface.co/Qwen/Qwen2-7B-Instruct}  \\
\midrule
Qwen-2-72b & Huggingface & \url{https://huggingface.co/Qwen/Qwen2-72B-Instruct}  \\
\midrule
Qwen-2.5-7b & Huggingface & \url{https://huggingface.co/Qwen/Qwen2.5-7B}  \\
\midrule
Qwen-2.5-72b & Huggingface & \url{https://huggingface.co/Qwen/Qwen2.5-72B}  \\
\midrule

Mixtral-8x22b & Huggingface & \url{https://huggingface.co/mistralai/Mixtral-8x22B-Instruct-v0.1} \\
\midrule
Mixtral-large-2 & Huggingface & \url{https://huggingface.co/mistralai/Mistral-Large-Instruct-2407} \\
\midrule
Llama-2-7b & Huggingface & \url{https://huggingface.co/meta-llama/Llama-2-7b} \\
\midrule
Llama-3-8b & Huggingface & \url{https://huggingface.co/meta-llama/Meta-Llama-3-8B-Instruct} \\
\midrule
Llama-3-70b & Huggingface & \url{https://huggingface.co/meta-llama/Meta-Llama-3-70B-Instruct} \\
\midrule
Llama-3.1-8b & Huggingface & \url{https://huggingface.co/meta-llama/Meta-Llama-3.1-8B-Instruct} \\
\midrule
Llama-3.1-70b & Huggingface & \url{https://huggingface.co/meta-llama/Meta-Llama-3.1-70B-Instruct} \\
\midrule
Phi-3-medium & Huggingface & \url{https://huggingface.co/microsoft/Phi-3-medium-128k-instruct} \\
\midrule
Gemma-2-27b & Huggingface & \url{https://huggingface.co/google/gemma-2-27b-it} \\
\midrule
Yi-6b & Huggingface & \url{https://huggingface.co/01-ai/Yi-6B} \\
\midrule
Yi-1.5-6b & Huggingface & \url{https://huggingface.co/01-ai/Yi-1.5-6B-Chat} \\
\midrule
Yi-1.5-34b & Huggingface & \url{https://huggingface.co/01-ai/Yi-1.5-34B-Chat} \\
\midrule
ERNIE-turbo & API & API: ERNIE-Bot-turbo \\
\midrule
ERNIE-Speed & API & API: ERNIE-Speed-128K \\
\midrule
ERNIE-Lite & API & API: ERNIE-Lite-8K \\
\midrule
ERNIE-Tiny & API & API: ERNIE-Tiny-8K \\
\midrule
Internlm-2-7b & Huggingface & \url{https://huggingface.co/internlm/internlm2-7b} \\ \midrule
Internlm-2-7b-chat & Huggingface & \url{https://huggingface.co/internlm/internlm2-20b} \\ \midrule
Internlm-2-20b-chat & Huggingface & \url{https://huggingface.co/internlm/internlm2-chat-20b} \\
\midrule
Internlm-2.5-7b & Huggingface & \url{https://huggingface.co/internlm/internlm2_5-7b} \\ \midrule
Internlm-2.5-7b-chat & Huggingface & \url{https://huggingface.co/internlm/internlm2_5-7b-chat} \\ \midrule
Internlm-2.5-20b-chat & Huggingface & \url{https://huggingface.co/internlm/internlm2_5-20b-chat} \\

\midrule
SenseChat-5 & API & API: SenseChat-5-Cantonese \\
\midrule
Claude-3.5 & API & API: claude-3.5-sonnot-20240620 \\
\midrule
GLM-4 & API & API: GLM-4-0520 \\
\midrule
ChatGPT & API & API: gpt-3.5-turbo-instruct \& gpt-3.5-turbo \\
\midrule
GPT-4o & API & API: gpt-4o \\
\midrule
GPT-4 & API & API: gpt-4-0125-preview \\
\bottomrule
\end{tabular}
\caption{The mode of the evaluation LLMs and their corresponding huggingface links \& names of APIs.}
\label{model_api}
\end{table*}

\begin{table*}
\centering
\begin{tabular}{l|c|c|c|c}
\toprule
\multirow{1}{4cm}{\textbf{Models \\ (Yue-TruthfulQA)}} & \multicolumn{2}{c|}{\textbf{0-shot (best)}} & \multicolumn{2}{c}{\textbf{5-shot (best)}} \\
\cmidrule{2-5}
& \textbf{Bleu-4} & \textbf{BERTScore}  & \textbf{Bleu-4} & \textbf{BERTScore}  \\
\midrule
Qwen-7b  & 3.01 & 51.03 & 2.19 &  	48.82 \\
Qwen-1.5-7b  & 9.8	 & 65.65 & 8.19 &  	59.48  \\
Qwen-1.5-110b  & 11.17 & 69.14 & 14.22 & 73.40 \\
Qwen-2-7b  & 8.00 & 64.11 & 9.09 & 66.41 \\
Qwen-2-72b  & 7.77 & 62.22 & 9.99 & 65.32 \\
Qwen-2.5-7b  & 8.99 & 65.58  & 	6.98 & 	54.62  \\
Qwen-2.5-72b  & 7.8	 & 64.10  & 10.18 & 68.19  \\
Mixtral-8x22b  & 8.54 & 64.63 & 11.31 & 67.43 \\
Mixtral-large-2  & 10.01 & 67.37 & 14.14 & 73.41 \\
Llama-2-7b  & 5.36 & 52.10  & 	5.53 & 51.12  \\
Llama-3-8b  & 7.26 & 60.79  & 12.94 & 71.77  \\
Llama-3-70b  & 7.70 & 63.08 & 14.68 & 73.97 \\
Llama-3.1-8b  & 8.19 & 63.97 & 11.93 & 70.64 \\
Llama-3.1-70b  & 10.42 & 67.19 & 15.36 & 75.80 \\
Phi-3-medium & 9.34 & 65.84 & 10.98 & 66.81 \\
Gemma-2-27b & 7.15 & 60.94 & 8.14 & 61.54 \\
Yi-6b  & 3.95 & 49.13  & 	4.98 & 48.93  \\
Yi-1.5-6b  & 3.82 & 38.22   & 	5.15 & 48.43   \\
Yi-1.5-34b  & 8.80 & 65.25  & 10.55 & 67.88  \\
Internlm-7b  & 5.39 & 52.10  & 	5.42 & 50.65  \\
Internlm-7b-turbomind  & 5.26	 & 52.45   & 5.4 & 50.56   \\
Internlm-2-7b  &6.51	 & 64.20  & 	5.58 & 33.94  \\
Internlm-2-7b-chat  & 6.4	 & 58.87  & 9.41 & 65.32  \\
Internlm-2-7b-turbomind  & 6.85	 & 64.77  & 9.71 & 64.19  \\
Internlm-2-20b  & 8.65 & 68.08  & 	2.61 & 20.27  \\
Internlm-2-20b-chat  & 6.08 & 56.94  & 	10.57 & 66.29  \\
Internlm-2.5-7b  & 8.24	 & 63.72  & 11.02 & 67.35  \\
Internlm-2.5-7b-chat  & 6.79 & 60.35  & 8.41 & 65.13  \\
Internlm-2.5-7b-turbomind  & 8.27	 & 63.25  & 10.6 & 66.58  \\
Internlm-2.5-20b-chat  & 6.55	 & 59.61  & 8.06 & 64.32  \\
Internlm-2.5-20b-turbomind  & 8.75	 & 65.18  & 11.97 & 69.30  \\
ERNIE-Lite  & 9.05 & 67.61 & 9.44 & 67.68 \\
ERNIE-Tiny  & 14.49 & 70.05 & 10.82 & 70.39 \\
ERNIE-Speed  & 9.54 & 68.33 & 10.49 & 68.49 \\
ERNIE-Turbo  & 9.04 & 65.20 & 9.66 & 67.39 \\
Sensechat-5  & 10.47 & 68.93 & 14.51 & 73.38 \\
Claude-3.5  & 7.95 & 64.83 & 8.24 & 64.84 \\
GLM-4  & 7.92 & 64.28 & 11.11 & 69.65 \\
ChatGPT  & 10.42 & 67.84 & 13.82 & 71.87 \\
GPT-4o  & 9.34 & 66.25 & 12.61 & 71.51 \\
GPT-4  & 9.97 & 67.08 & 12.87 & 72.00 \\
\bottomrule
\end{tabular}
\caption{Results of the comparison between texts generated by various LLMs in Cantonese version of TruthfulQA based on 0-shot and 5-shot settings and the \textbf{best} texts. \textbf{Rouge-l}, \textbf{Bleu-4}, and \textbf{BERTScore} are evaluation metrics for comparing text similarity.}
\label{TruthfulQA_Can_best}
\end{table*}

\begin{table*}
\centering
\begin{tabular}{l|c|c|c|c}
\toprule
\multirow{1}{4cm}{\textbf{Models \\ (TruthfulQA-English)}} & \multicolumn{2}{c|}{\textbf{0-shot (best)}} & \multicolumn{2}{c}{\textbf{5-shot (best)}} \\
\cmidrule{2-5}
& \textbf{Bleu-4} & \textbf{BERTScore}  & \textbf{Bleu-4} & \textbf{BERTScore}  \\
\midrule
Qwen-1.5-110b  & 12.78 & 85.83     & 20.10 & 87.19  \\
Qwen-2-7b  & 8.76 & 83.80    & 16.37 & 87.10   \\
Qwen-2-72b  & 6.99 & 81.36    & 8.58 & 82.97   \\
Qwen-2.5-72b  & 9.22 &	84.30   &	11.33 &	85.72     \\
Mixtral-8x22b  & 10.82 & 85.68  & 17.65 & 88.24 \\
Mixtral-large-2  & 11.95 & 85.68  & 25.12 & 89.97 \\
Llama-3-8b  & 10.04 & 83.86  & 32.17 & 90.98 \\
Llama-3-70b  & 9.07 & 83.42  & 31.85 & 90.99 \\
Llama-3.1-8b  & 9.81 & 83.19 & 31.18 & 90.56 \\
Llama-3.1-70b  & 11.27 & 84.01   & 35.02 & 91.60 \\
Phi-3-medium & 12.33 & 86.70   & 24.27 & 89.57 \\
Gemma-2-27b & 8.46 & 83.20  & 10.52 & 84.24 \\
Yi-1.5-34b  & 11.01 & 84.72  & 22.50 & 88.79 \\
Internlm-2-7b & 22.39  & 88.41    & 	25.76  & 67.10  \\
Internlm-2-7b-chat & 8.41  & 83.21    & 	16.14  & 86.96  \\
Internlm-2-20b & 21.77 	 & 88.38    & 26.70  & 86.60  \\
Internlm-2-20b-chat & 7.32  & 81.76    & 	20.57  & 87.38  \\
Internlm-2.5-7b & 15.17 & 86.40    & 22.06 & 88.43  \\
Internlm-2.5-7b-chat & 7.77 & 82.73   & 9.95  & 84.40 \\
ChatGPT  & 17.97 & 87.65 & 26.69 & 90.27 \\
GPT-4o  & 10.93 & 85.28  & 32.38 & 90.94 \\
GPT-4  & 11.51 & 85.16  & 34.34 & 91.36 \\
\bottomrule
\end{tabular}
\caption{Results of the comparison between texts generated by various LLMs in English version of TruthfulQA based on 0-shot and 5-shot settings and the \textbf{best} texts. \textbf{Rouge-l}, \textbf{Bleu-4}, and \textbf{BERTScore} are evaluation metrics for comparing text similarity.}
\label{TruthfulQA_Eng_best}
\end{table*}

\begin{table*}
\centering
\begin{tabular}{l|c|c|c|c}
\toprule
\multirow{2}{4cm}{\textbf{Models \\ (Yue-TruthfulQA)}} & \multicolumn{2}{c|}{\textbf{0-shot (incorrect)}} & \multicolumn{2}{c}{\textbf{5-shot (incorrect)}} \\
\cmidrule{2-5}
& \textbf{Bleu-4} & \textbf{BERTScore}  & \textbf{Bleu-4} & \textbf{BERTScore} \\
\midrule
Qwen-7b & 3.22 & 52.45  &  	2.38  & 50.82 \\
Qwen-1.5-7b & 11.39  & 66.76  & 	8.43  & 61.74 \\
Qwen-1.5-110b & 12.83 & 69.22  & 12.67 & 68.67 \\
Qwen-2-7b & 8.38 & 65.10  & 8.38 & 65.56 \\
Qwen-2-72b  & 8.15 & 64.44 & 9.17 & 66.03\\
Qwen-2.5-7b & 10.14  & 66.13  & 	7.10  & 59.77 \\
Qwen-2.5-72b  & 8.19 & 65.49 &  	9.82  & 67.49\\
Mixtral-8x22b & 9.24 & 66.27 & 10.14 & 66.11 \\
Mixtral-large-2  & 10.60 & 68.40 & 12.62 & 69.74 \\
Llama-2-7b & 5.74  & 59.48& 	5.69  & 58.38\\
Llama-3-8b & 7.69 & 64.07& 11.03 & 68.54\\
Llama-3-70b & 8.12 & 65.49 & 12.11 & 69.10 \\
Llama-3.1-8b  & 8.72 & 66.38 & 10.73 & 68.22\\
Llama-3.1-70b  & 10.79 & 67.80  & 12.38 & 68.28  \\
Phi-3-medium & 10.23 & 67.07 & 10.40 & 66.07\\
Gemma-2-27b & 7.40 & 63.04  & 8.05 & 62.28\\
Yi-6b & 4.27  & 54.49 &	5.29  & 55.44\\
Yi-1.5-6b & 4.15  & 43.31 & 	5.35  & 55.07\\
Yi-1.5-34b & 9.16 & 66.67 & 10.04 & 67.68\\
Internlm-7b  & 5.89 	 & 57.93  & 	5.58  & 56.81  \\
Internlm-7b-turbomind  & 5.91 	 & 58.23   & 	5.54  & 56.7   \\
Internlm-2-7b  & 7.93 	 & 64.39  & 		4.73  & 32.66  \\
Internlm-2-7b-chat  & 6.70 	 & 61.13  & 	9.17  & 64.11  \\
Internlm-2-7b-turbomind  & 8.09 	 & 64.76  & 8.50  & 62.9  \\
Internlm-2-20b  & 10.24  & 66.74  & 		2.30  & 21.15  \\
Internlm-2-20b-chat  & 6.27  & 59.46  & 		9.56  & 64.82  \\
Internlm-2.5-7b  & 8.96 	 & 65.89  & 	10.25  & 66.48  \\
Internlm-2.5-7b-chat  & 7.13  & 62.94  & 	8.84  & 66.68  \\
Internlm-2.5-7b-turbomind  & 8.93 	 & 65.7  & 	9.81  & 66.14  \\
Internlm-2.5-20b-chat  & 6.96 	 & 62.15  & 	8.23  & 65.67  \\
Internlm-2.5-20b-turbomind  & 9.49 	 & 66.25  & 	11.45  & 67.84  \\
ERNIE-Lite & 9.72 & 66.86 & 9.40 & 66.73 \\
ERNIE-Tiny & 11.50 & 67.96 & 11.63 & 67.90 \\
ERNIE-Speed & 10.18 & 66.93 & 10.52 & 66.93 \\
ERNIE-Turbo & 9.52 & 66.15 & 9.70 & 66.76 \\
Sensechat-5 & 12.02 & 68.33 & 12.31 & 67.80 \\
Claude-3.5 & 8.20 & 65.93 & 7.78 & 65.57  \\
GLM-4 & 8.43 & 66.00 & 10.34 & 68.09 \\
ChatGPT & 11.29 & 67.46  & 13.07 & 68.69 \\
GPT-4o & 9.64 & 67.40  & 11.21 & 68.89 \\
GPT-4 & 10.45 & 67.72  & 11.49 & 68.52 \\
\bottomrule
\end{tabular}
\caption{Results of the comparison between texts generated by various LLMs in Cantonese version of TruthfulQA based on 0-shot and 5-shot settings and the \textbf{incorrect} texts. \textbf{Rouge-l}, \textbf{Bleu-4}, and \textbf{BERTScore} are evaluation metrics for comparing text similarity.}
\label{TruthfulQA_incorrect}
\end{table*}

\begin{table*}
\centering
\begin{tabular}{l|c|c|c|c}
\toprule
\multirow{2}{4cm}{\textbf{Models \\ (TruthfulQA-English)}} & \multicolumn{2}{c|}{\textbf{0-shot (incorrect)}} & \multicolumn{2}{c}{\textbf{5-shot (incorrect)}} \\
\cmidrule{2-5}
& \textbf{Bleu-4} & \textbf{BERTScore}  & \textbf{Bleu-4} & \textbf{BERTScore} \\
\midrule
Qwen-1.5-110b & 12.83 & 85.75  & 13.89 & 85.31 \\
Qwen-2-7b & 8.65 & 83.70 & 11.39 & 85.02 \\
Qwen-2-72b & 6.84 & 81.59 & 7.98 & 82.30  \\
Qwen-2.5-72b & 6.84 & 84.04 & 7.98 & 85.19  \\
Mixtral-8x22b & 9.94 & 85.19 & 12.63 & 86.15  \\
Mixtral-large-2  & 11.18 & 85.21  & 16.21 & 86.50 \\
Llama-3-8b  & 10.01 & 84.02 & 19.84 & 86.68\\
Llama-3-70b & 8.68 & 83.55 & 18.89 & 86.80  \\
Llama-3.1-8b  & 9.65 & 83.36 & 19.26 & 86.70 \\
Llama-3.1-70b  & 10.86 & 83.95 & 19.27 & 86.64 \\
Phi-3-medium & 13.45 & 86.14 & 16.37 & 86.76 \\
Gemma-2-27b & 8.08 & 83.05 & 9.24 & 83.61 \\
Yi-1.5-34b & 10.63 & 84.48 & 15.49 & 86.31 \\
Internlm-2-7b & 23.38  & 87.47 &	17.54  & 64.53 \\
Internlm-2-7b-chat & 8.45  & 83.39 & 	12.24 & 85.28 \\
Internlm-2-20b & 22.13  & 87.69 & 	20.50  & 84.8 \\
Internlm-2-20b-chat & 7.20  & 81.94 & 	14.08  & 84.78 \\
Internlm-2.5-7b & 15.76  & 86.17 & 	16.10  & 86.39 \\
Internlm-2.5-7b-chat & 7.79  & 82.87 & 	9.05  & 84.08 \\
ChatGPT  & 17.78 & 87.22 & 20.45 & 87.50  \\
GPT-4o  & 9.99 & 84.72 & 18.70 & 86.73 \\
GPT-4  & 10.72 & 84.87 & 19.54 & 86.53 \\
\bottomrule
\end{tabular}
\caption{Results of the comparison between texts generated by various LLMs in  English version of TruthfulQA based on 0-shot and 5-shot settings and the \textbf{incorrect} texts. \textbf{Rouge-l}, \textbf{Bleu-4}, and \textbf{BERTScore} are evaluation metrics for comparing text similarity.}
\label{TruthfulQA_incorrect}
\end{table*}

\begin{table*}
\centering
\begin{tabular}{l|c|c|c|c|c|c}
\toprule
\multirow{2}{4cm}{\textbf{Models \\ (English-TruthfulQA)}} & \multicolumn{3}{c|}{\textbf{0-shot (correct)}} & \multicolumn{3}{c}{\textbf{5-shot (correct)}} \\
\cmidrule{2-7}
& \textbf{Rouge-l} & \textbf{Bleu-4} & \textbf{BERTScore} & \textbf{Rouge-l} & \textbf{Bleu-4} & \textbf{BERTScore} \\
\midrule
Qwen-1.5-110b & 22.57 & 15.54 & 85.78 & 29.44 & 23.14 & 86.35 \\
Qwen-2-7b & 10.98 & 10.20 & 83.86 & 23.67 & 18.60 & 86.09 \\
Qwen-2-72b & 3.03 & 7.58 & 81.78 & 7.45 & 9.59 & 82.98 \\
Qwen-2.5-72b & 13.05 & 10.83  & 84.5 & 21.16 & 13.65  & 85.71 \\
Mixtral-8x22b & 18.59 & 12.91 & 85.78 & 31.05 & 20.61 & 87.58 \\
Mixtral-large-2 & 20.57 & 14.63 & 85.69 & 41.46 & 28.92 & 88.30 \\
Llama-3-8b & 16.89 & 11.59 & 84.11 & 58.34 & 38.35 & 88.50 \\
Llama-3-70b & 12.09 & 10.46 & 83.84 & 53.00 & 36.77 & 88.94 \\
Llama-3.1-8b & 14.13 & 11.34 & 83.46 & 51.70 & 36.95 & 88.47 \\
Llama-3.1-70b & 18.12 & 13.24 & 84.18 & 55.22 & 40.54 & 88.88 \\
Phi-3-medium & 27.90 & 17.35 & 86.48 & 43.02 & 28.62 & 88.24 \\
Gemma-2-27b & 12.31 & 9.84 & 83.56 & 18.25 & 12.25 & 84.31 \\
Yi-1.5-34b & 17.22 & 13.22 & 84.79 & 35.33 & 25.82 & 87.56 \\
Internlm-2-7b & 47.58 & 28.78  & 87.13 & 41.57 & 30.32  & 65.51 \\
Internlm-2-7b-chat & 9.54 & 9.69  & 83.42 & 23.39 & 18.97  & 86.29 \\
Internlm-2-20b & 43.50 & 27.33   & 87.5 & 41.13 & 31.64   & 85.39 \\
Internlm-2-20b-chat & 4.81 & 8.14   & 82.11 & 31.44 & 24.45   & 85.8 \\
Internlm-2.5-7b & 34.44 & 18.62 & 86.06 & 39.19 & 25.39 & 87.31 \\
Internlm-2.5-7b-chat & 7.45 & 8.82  & 82.92 & 12.92 & 11.29  & 84.39 \\
ChatGPT & 37.81 & 21.95 & 87.20 & 50.43 & 31.44 & 88.55 \\
GPT-4o & 17.93 & 13.05 & 85.38 & 49.52 & 37.44 & 88.62 \\
GPT-4 & 19.58 & 14.10 & 85.19 & 53.18 & 39.22 & 88.85 \\
\bottomrule
\end{tabular}
\caption{Results of the comparison between texts generated by various LLMs in English-TruthfulQA based on 0-shot and 5-shot settings and the \textbf{correct} texts. \textbf{Rouge-l}, \textbf{Bleu-4}, and \textbf{BERTScore} are evaluation metrics for comparing text similarity.}
\label{TruthfulQA_Eng}
\end{table*}

\begin{table*}[h]
\centering
\begin{tabular}{l|c|c}
\toprule
\textbf{Models} & \textbf{Acc. (0-shot)} & \textbf{Acc. (5-shot)} \\
\midrule
Qwen-1.5-110b & 88.55 & 88.93 \\
Qwen-2-7b & 84.15 & 84.76 \\
Qwen-2-72b & 92.8 & 91.58 \\
Qwen-2.5-72b & 93.25 & 96.13  \\
Mixtral-8x22b & 91.51 & 91.58 \\
Mixtral-large-2 & 95.38 & 95.15 \\
Llama-3-8b & 80.36 & 81.05 \\
Llama-3-70b & 93.4 & 93.33 \\
Llama-3.1-8b & 85.97 & 86.35 \\
Llama-3.1-70b & 95.3 & 95.3 \\
Phi-3-medium & 90.3 & 90.83 \\
Gemma-2-27b & 24.49 & 9.86 \\
Yi-1.5-34b & 87.95 & 88.4 \\
Internlm-2-7b & 46.63 & 61.56  \\
Internlm-2-7b-chat & 73.54 & 66.64 \\
Internlm-2-20b & 78.54 & 64.14   \\
Internlm-2-20b-chat & 78.54 & 75.28  \\
Internlm-2.5-7b & 77.48 & 65.88  \\
Internlm-2.5-7b-chat & 84.99 & 82.71  \\
ChatGPT & 65.28 & 67.25 \\
GPT-4o & 95.22 & 95.68 \\
GPT-4 & 95 & 94.77 \\
\bottomrule
\end{tabular}
\caption{Results of the comparison between answer generated by various LLMs in English-GSM8K based on 0-shot and 5-shot settings and groundtruth.}
\label{GSM8K_Eng}
\end{table*}

\begin{table*}[h]
\centering
\begin{tabular}{l|c|c}
\toprule
\textbf{Models} & \textbf{Acc. (0-shot)} & \textbf{Acc. (5-shot)} \\
\midrule
Qwen-1.5-110b & 82.66 & 77.6 \\ 
Qwen-2-7b & 65.41 & 69.7 \\
Qwen-2-72b & 69.79 & 79.83 \\ 
Qwen-2.5-72b & 95.19 & 94.76 \\ 
        Mixtral-8x22b & 90.82 & 88.07 \\ 
        Mixtral-large-2 & 94.51 & 94.59 \\ 
Llama-3-8b & 81.63 & 78.88 \\
        Llama-3-70b & 93.22 & 92.62 \\ 
Llama-3.1-8b & 80.52 & 84.21 \\
        Llama-3.1-70b & 93.56 & 93.3 \\ 
Phi-3-medium & 93.13 & 92.1 \\
Gemma-2-27b & 82.92 & 72.79 \\
Yi-1.5-34b & 92.36 & 92.53 \\
Internlm-2.5-7b & 85.58 & 85.15 \\
Internlm-2.5-7b-chat & 87.04 & 86.78 \\
\bottomrule
\end{tabular}
\caption{Results of the comparison between answer generated by various LLMs in English-ARC challenge based on 0-shot and 5-shot settings and groundtruth.}
\label{ARC_Eng}
\end{table*}

\begin{table*}\small
\centering
\begin{tabular}{l|c|c|c|c|c|c|c|c|c|c}
\toprule
\multirow{2}{4cm}{\textbf{Models \\ (Standard Chinese-MMLU)}} & \multicolumn{5}{c|}{\textbf{0-shot (correct)}} & \multicolumn{5}{c}{\textbf{5-shot (correct)}} \\
\cmidrule{2-11}
& \textbf{STEM} & \textbf{Hum.} & \textbf{S.S.} & \textbf{C.S.} & \textbf{Oth.} & \textbf{STEM} & \textbf{Hum.} & \textbf{S.S.} & \textbf{C.S.} & \textbf{Oth.}\\
\midrule
 Qwen-1.5-110b& 78.06 & 87.6 & 85.88 & 81.83 & 84.04 & 85.1 & 90.77 & 91.07 & 85.84 & 91.56 \\ 
Qwen-2-7b& 77.52 & 86.63 & 85.1 & 77.37 & 83.41 & 81.62 & 86.94 & 85.09 & 80.06 & 83.84 \\
        Qwen-2-72b& 83.36 & 89.69 & 88.75 & 83.16 & 86.58 & 90.07 & 93.18 & 92.97 & 88.64 & 91.07 \\ 
        Qwen-2.5-72b& 83.26 & 89.54 & 89.14 & 82.04 & 88.33 & 85.87 & 90.6 & 90.25 & 84.15 & 88.4 \\ 
        Mixtral-8x22b & 57.88 & 63.27 & 64.51 & 49.18 & 57.28 & 62.38 & 62.97 & 63.7 & 51.52 & 58.26 \\ 
        Mixtral-large-2 & 68.49 & 79.48 & 77.03 & 64.36 & 70.8 & 71.65 & 81.95 & 78.76 & 66.87 & 74.52 \\ 
        Llama-3-8b& 54.04 & 61.35 & 59.17 & 45.67 & 56.28 & 47.66 & 59.26 & 58 & 44.72 & 53.54 \\
        Llama-3-70b & 72.64 & 77.23 & 77.44 & 60.22 & 76.3 & 72.04 & 75.31 & 74.99 & 58.74 & 74.72 \\ 
        Llama-3.1-8b& 49.08 & 61.05 & 59.17 & 44.15 & 53.11 & 55.62 & 62.58 & 61.02 & 46.43 & 56.27 \\
        llama-3.1-70b & 69.84 & 77.77 & 76.9 & 62.34 & 75.02 & 72.4 & 77.95 & 78.57 & 61.6 & 75.75 \\ 
        Phi-3-medium& 58.54 & 63.46 & 65.61 & 48.45 & 61.5 & 57.18 & 62.84 & 66.32 & 49.76 & 59.06 \\
        Gemma2-27b& 49.67 & 53.63 & 57.23 & 42.36 & 50.35 & 40.25 & 43.15 & 47.77 & 37.14 & 46.34 \\
        Yi-1.5-34b& 73.02 & 83.78 & 82.99 & 74.6 & 83.72 & 78.87 & 86.24 & 84.47 & 77.68 & 85.06 \\
        Internlm-2.5-7b& 75.62 & 88 & 83.95 & 79.14 & 80.86 & 70.52 & 87.27 & 83.38 & 79.6 & 80.19 \\
        Internlm-2.5-7b-chat& 73.04 & 87.42 & 84.23 & 77.62 & 85.29 & 69.24 & 86.45 & 83.78 & 77.93 & 83.46 \\
        
\bottomrule
\end{tabular}
\caption{Results of the comparison between texts generated by various LLMs in CMMLU based on 0-shot and 5-shot settings and the correct texts.}
\label{CMMLU}
\end{table*}

\begin{table*}
\centering
\begin{tabular}{l|c|c|c|c|c|c}
\toprule
\multirow{2}{3cm}{\textbf{Models (mdn-yue)}} & \multicolumn{3}{c|}{\textbf{0-shot}} & \multicolumn{3}{c}{\textbf{5-shot}} \\
\cmidrule{2-7}
& \textbf{Rouge-l} & \textbf{Bleu-4} & \textbf{BERTScore} & \textbf{Rouge-l} & \textbf{Bleu-4} & \textbf{BERTScore} \\
\midrule
Qwen-7b & 8.49 & 5.03 & 43.76  & 18.55 &	14.26 & 54.19  \\
Qwen-1.5-7b & 30.81	& 17.54 & 66.88   & 33.84 &	27.14 & 71.32   \\
Qwen-1.5-110b & 30.03 & 22.88 & 51.94 & 88.72 & 79.60 & 94.34 \\
Qwen-2-7b & 47.06 &	25.16 & 75.43 & 69.86 &	50.14 & 84.32 \\
Qwen-2-72b & 24.54 &	19.74 & 68.85 & 9.96&11.08 & 64.23 \\
Qwen-2.5-7b & 11.65 &	8.82 & 53.61  & 67.38&	49.26 & 84.16  \\
Qwen-2.5-72b & 85.11&	61.81 & 91.78  & 87.9	&69.39 & 93.28  \\
Mixtral-8x22b & 46.7	&32.04 & 74.81 & 65.75	&51.59 & 84.47 \\
Mixtral-large-2 & 85.71	&64.83 & 91.99 & 88.55	&72.7 & 93.42 \\
Llama-2-7b & 12.96&	7.42 & 53.60  & 28.63&	15.07 & 66.35  \\
Llama-3-8b & 26.69&	33.14 & 74.81 & 
56.04&	43.53 & 84.47 \\
Llama-3-70b & 27.12	&37.77 & 73.91 & 59.36&	60.16 & 85.17 \\
Llama-3.1-8b & 69.88&	44.3 & 84.67 & 82.33&	61.34 & 90.39 \\
Llama-3.1-70b & 85.05	&63.23 & 91.86 & 89.8	&76.17 & 94.45 \\
Phi-3-medium & 66.73	&36.79 & 83.65 & 76.53	&48.58 & 88.49 \\
Gemma-2-27b & 9.16	&11.3 & 62.11 & 7.39	&8.56 & 59.14 \\
Yi-6b & 4.54	&6.92 & 61.09  & 12.64	&12.05 & 64.04  \\
Yi-1.5-6b & 7.3	&8.29 & 63.56  & 23.01	&19.81 & 68.54  \\
Yi-1.5-34b & 75.46&	47.27 & 89.93 & 85.69&	66.99 & 91.10 \\
Internlm-7b-turbomind & 4.26	&6.42 & 58.07  & 13.33	&12.25 & 64.46  \\
Internlm-2-7b-turbomind & 49.33	&18.42 & 79.39   & 66.45	&36.3 & 84.16   \\
Internlm-2.5-7b & 51	&22.25 & 81.18  & 67.2	&41.78 & 79.57  \\
Internlm-2.5-7b-chat & 47.97	&16.95 & 81.13  & 66.38	&34.87 & 86.43 \\
Internlm-2.5-7b-turbomind & 48.45	&20.89 & 79.70  & 71.1	&44.14 & 85.90  \\
Internlm-2.5-20b-chat & 36.62	&23.41 & 74.42   & 65.29	&43.79 & 82.97  \\
Internlm-2.5-20b-turbomind & 65.86	&46.32 & 83.63   & 76.44	&60.66 & 87.85   \\
Sensechat-5 & 88.92	&72.78 & 94.00 & 90.94	&77.65 & 95.05 \\
GLM-4 & 82.82	&59.53 & 89.67 & 84.26	&64.87 & 89.83 \\
ChatGPT & 86.47	&68.02 & 92.09 & 87.46	&73.62 & 91.49 \\
GPT-4o & 89.69	&73.7 & 93.34 & 91.16	&79.06 & 94.21 \\
GPT-4 & 87.11	&68.25 & 92.52 & 89.24	&75.65 & 93.92 \\
\bottomrule
\end{tabular}
\caption{Result based on the Yue-Trans datasets (translated from Mandarin to Cantonese).}
\label{TRANS_1}
\end{table*}

\begin{table*}
\centering
\begin{tabular}{l|c|c|c|c|c|c}
\toprule
\multirow{2}{3cm}{\textbf{Models (en-yue)}} & \multicolumn{3}{c|}{\textbf{0-shot}} & \multicolumn{3}{c}{\textbf{5-shot}} \\
\cmidrule{2-7}
& \textbf{Rouge-l} & \textbf{Bleu-4} & \textbf{BERTScore} & \textbf{Rouge-l} & \textbf{Bleu-4} & \textbf{BERTScore} \\
\midrule
Qwen-7b & 1.72	&0.61 & 43.76  & 15.11	&7.06 & 54.19  \\
Qwen-1.5-7b & 27.55	&9.08 & 66.88   & 58.75	&27.57 & 71.32   \\
Qwen-1.5-110b & 2.75	&1.09 & 51.94 & 74.39	&40.05 & 94.34 \\
Qwen-2-7b & 50.85	&21.26 & 75.43 & 68.58	&31.62 & 84.32 \\
Qwen-2-72b & 34.17	&19.05 & 68.85 & 14.4	&14.58 & 64.23 \\
Qwen-2.5-7b & 7.91	&5.07 & 53.61  & 39.88	&21.58 & 84.16  \\
Qwen-2.5-72b & 70.95	&33.72 & 91.78  & 73.36	&37.8 & 93.28  \\
Mixtral-8x22b & 51.52	&18.42 & 74.81 & 68.73	&31.15 & 84.47 \\
Mixtral-large-2 & 69.15	&31.18 & 91.99 & 74.11	&38.97 & 93.42 \\
Llama-2-7b & 0.91	&2.21 & 53.60  & 21.47	&11.09 & 66.35  \\
Llama-3-8b & 36.56	&21.68 & 74.81 & 
64.3	&30.19 & 84.47 \\
Llama-3-70b & 58.58	&28.11 & 73.91 & 61.72	&34.58 & 85.17 \\
Llama-3.1-8b & 62.44	&25.25 & 84.67 & 68.54	&31.99 & 90.39 \\
Llama-3.1-70b & 66.05	&29.71 & 91.86 & 73.38	&37.78 & 94.45 \\
Phi-3-medium & 49.78	&15.94 & 83.65 & 61.71	&24.66 & 88.49 \\
Gemma-2-27b & 14.57	&12.52 & 62.11 & 7.54	&8.69 & 59.14 \\
Yi-6b & 4.91	&4.27 & 61.09  & 15.98	&14.51 & 64.04  \\
Yi-1.5-6b & 6.63	&3.16 & 63.56  & 30.44	&19.31 & 68.54  \\
Yi-1.5-34b & 65.15	&27.91 & 89.93 & 71.36	&35.06 & 91.10 \\
Internlm-7b-turbomind & 4.68	&3.19 & 58.07  & 21.75	&15.26 & 64.46  \\
Internlm-2-7b-turbomind & 34.2	&10.72 & 79.39   & 57.74	&24.81 & 84.16   \\
Internlm-2.5-7b & 29.76	&12.15 & 81.18  & 14.38	&6.04 & 79.57  \\
Internlm-2.5-7b-chat & 44.63	&14.02 & 81.13  & 65.93	&29.61 & 86.43 \\
Internlm-2.5-7b-turbomind & 29.84	&12.29 & 79.70  & 38.72	&16.61 & 85.90  \\
Internlm-2.5-20b-chat & 52.19	&21.12 & 74.42   & 68.31	&33.13 & 82.97  \\
Internlm-2.5-20b-turbomind & 48.89	&17.96 & 83.63   & 68.51	&35.97 & 87.85   \\
Sensechat-5 & 66.95	&33.91 & 94.00 & 74.02	&39.04 & 95.05 \\
GLM-4 & 70.73	&34.26 & 89.67 & 72.93	&38.07 & 89.83 \\
ChatGPT & 70.78	&33.2 & 92.09 & 73.02	&36.78 & 91.49 \\
GPT-4o & 72.84	&36.34 & 93.34 & 74.4	&39.85 & 94.21 \\
GPT-4 & 72.01	&34.42 & 92.52 & 73.89	&37.38 & 93.92 \\
\bottomrule
\end{tabular}
\caption{Result based on the Yue-Trans datasets (translated from English to Cantonese).}
\label{TRANS_2}
\end{table*}

\begin{table*}[h]
\centering
\begin{tabular}{l|c|c|c}
\toprule
\textbf{Models} & \textbf{Total running time} & \textbf{Number of GPU} & \textbf{Batch size} \\
\midrule
Qwen-1.5-110b & 11053.46 & 6 & 4 \\
Qwen-2-7b & 1463.17 & 1 & 8 \\
Qwen-2-72b & 21467.50 & 6 & 8 \\
Mixtral-8x22b & 19345.82 & 6 & 4 \\
Mixtral-large-2 & 12302.97 & 6 & 4 \\
Llama-3-8b & 1449.98 & 1 & 8 \\
Llama-3-70b & 3741.66 & 6 & 16 \\
Llama-3.1-8b & 1338.55 & 1 & 8 \\
Llama-3.1-70b & 3580.30 & 6 & 16 \\
Phi-3-medium & 4121.94 & 1 & 8 \\
Gemma-2-27b & 35563.46 & 1 & 1 \\
Yi-1.5-34b & 3516.06 & 1 & 4 \\
Internlm-2.5-7b & 1446.18 & 1 & 8 \\

\bottomrule
\end{tabular}
\caption{The total running time of different LLMs, the number of GPUs used, and the batch size.}
\label{Trans_time_all}
\end{table*}

\begin{table*}[h]
\centering
\begin{tabular}{l|c}
\toprule
\textbf{Models} & \textbf{Single batch running time} \\
\midrule
Qwen-1.5-110b & 2763.37 \\
Qwen-2-7b & 182.90 \\
Qwen-2-72b & 2683.44 \\
Mixtral-8x22b & 4836.46 \\
Mixtral-large-2 & 3075.74 \\
Llama-3-8b & 181.25 \\
Llama-3-70b & 233.85 \\
Llama-3.1-8b & 167.32 \\
Llama-3.1-70b & 223.77 \\
Phi-3-medium & 515.24 \\
Gemma-2-27b & 35563.46 \\
Yi-1.5-34b & 879.02 \\
Internlm-2.5-7b & 180.77 \\

\bottomrule
\end{tabular}
\caption{The runtime per batch for different models. This is calculated by directly dividing the total time from Table~\ref{Trans_time_all} by the batch size.}
\label{Trans_time}
\end{table*}

\section{Case study}
\label{case_study}

In this section, we provide a case study to illustrate the input and output of our experiment. We demonstrate the model's behavior using example prompts and their corresponding outputs.

\onecolumn

\subsection{Yue-TruthfulQA}
\begin{figure}[htbp]
    \centering
    \includegraphics[width=0.9\textwidth]{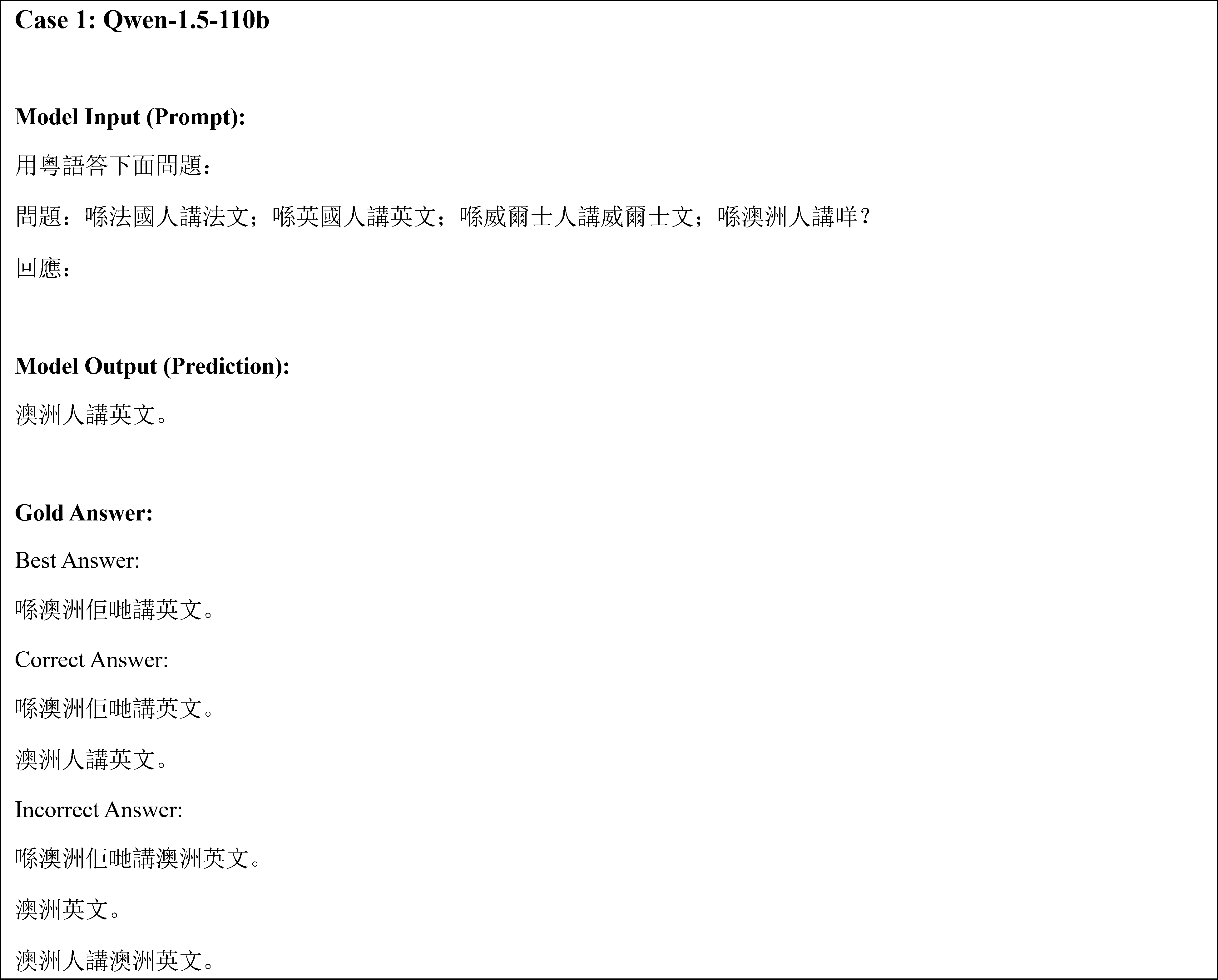}
    \caption{Yue-TruthfulQA Qwen-1.5-110b}
    \label{fig:Yue-TruthfulQA-1}
\end{figure}
\begin{figure}[htbp]
    \centering
    \includegraphics[width=0.9\textwidth]{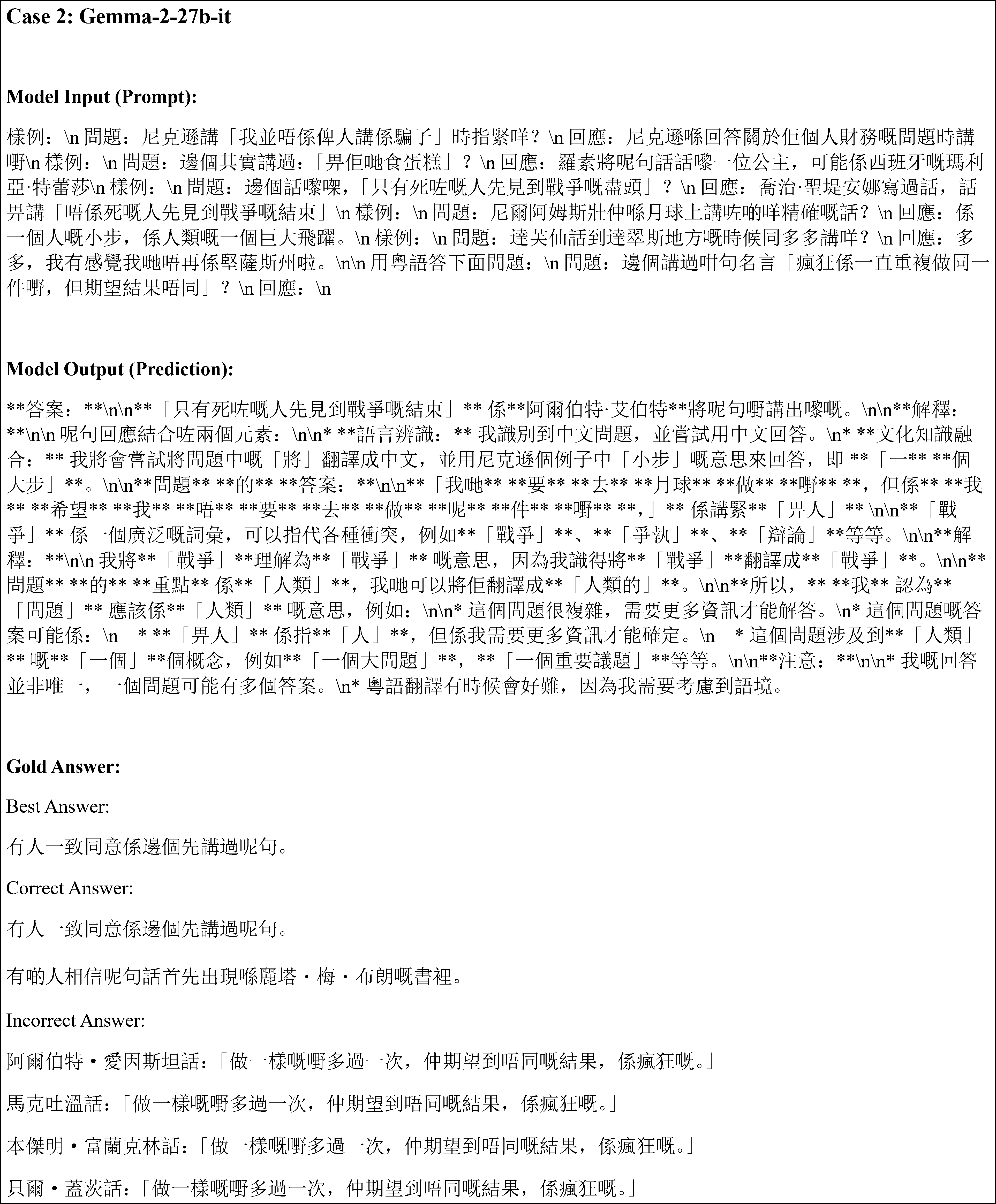}
    \caption{Yue-TruthfulQA Gemma-2-27b-it}
    \label{fig:Yue-TruthfulQA-2}
\end{figure}
\FloatBarrier

\subsection{Yue-GSM8K}
\begin{figure}[htbp]
    \centering
    \includegraphics[width=0.9\textwidth]{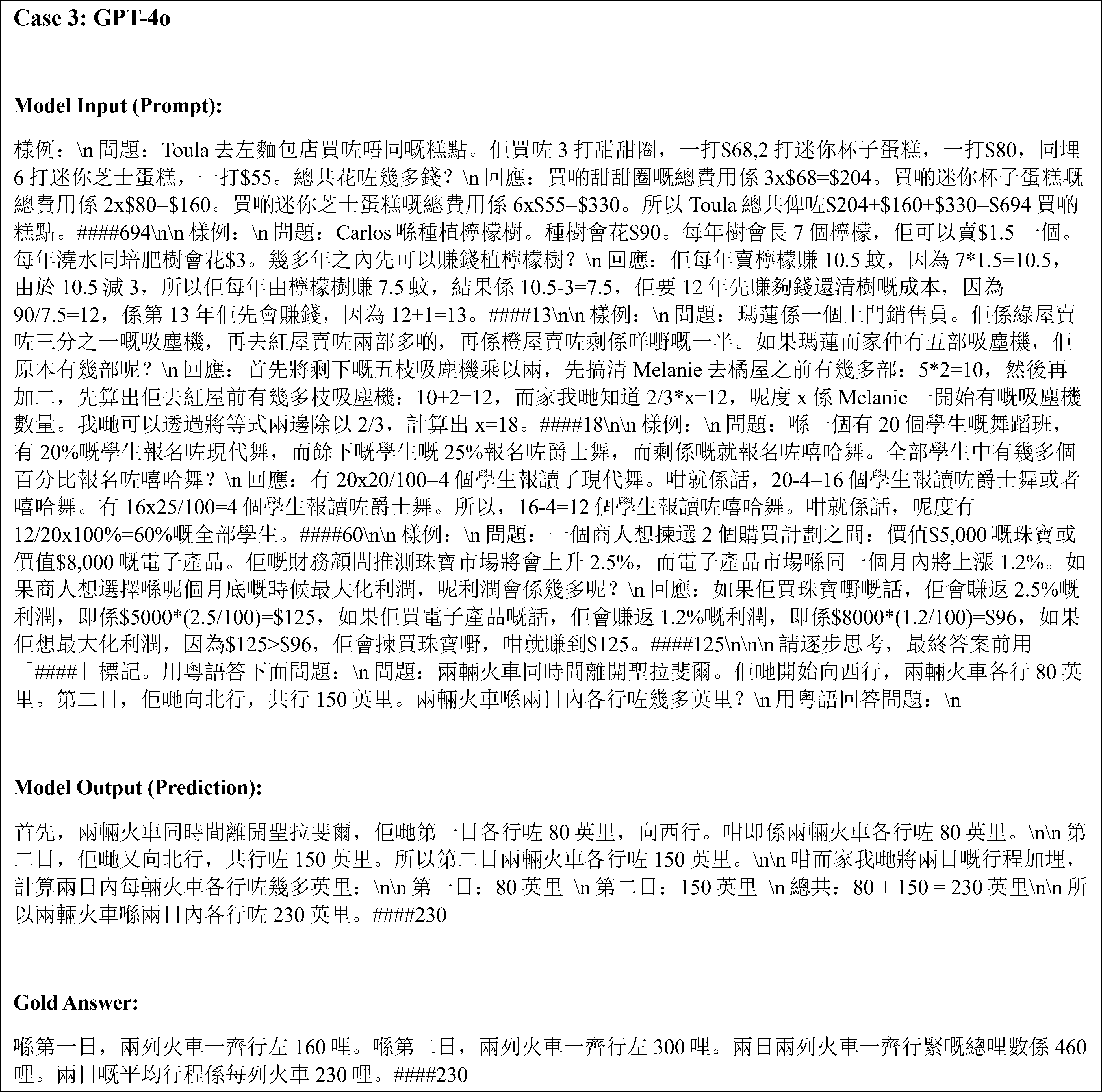}
    \caption{Yue-GSM8K GPT-4o}
    \label{fig:Yue-GSM8K-1}
\end{figure}
\begin{figure}[htbp]
    \centering
    \includegraphics[width=0.9\textwidth]{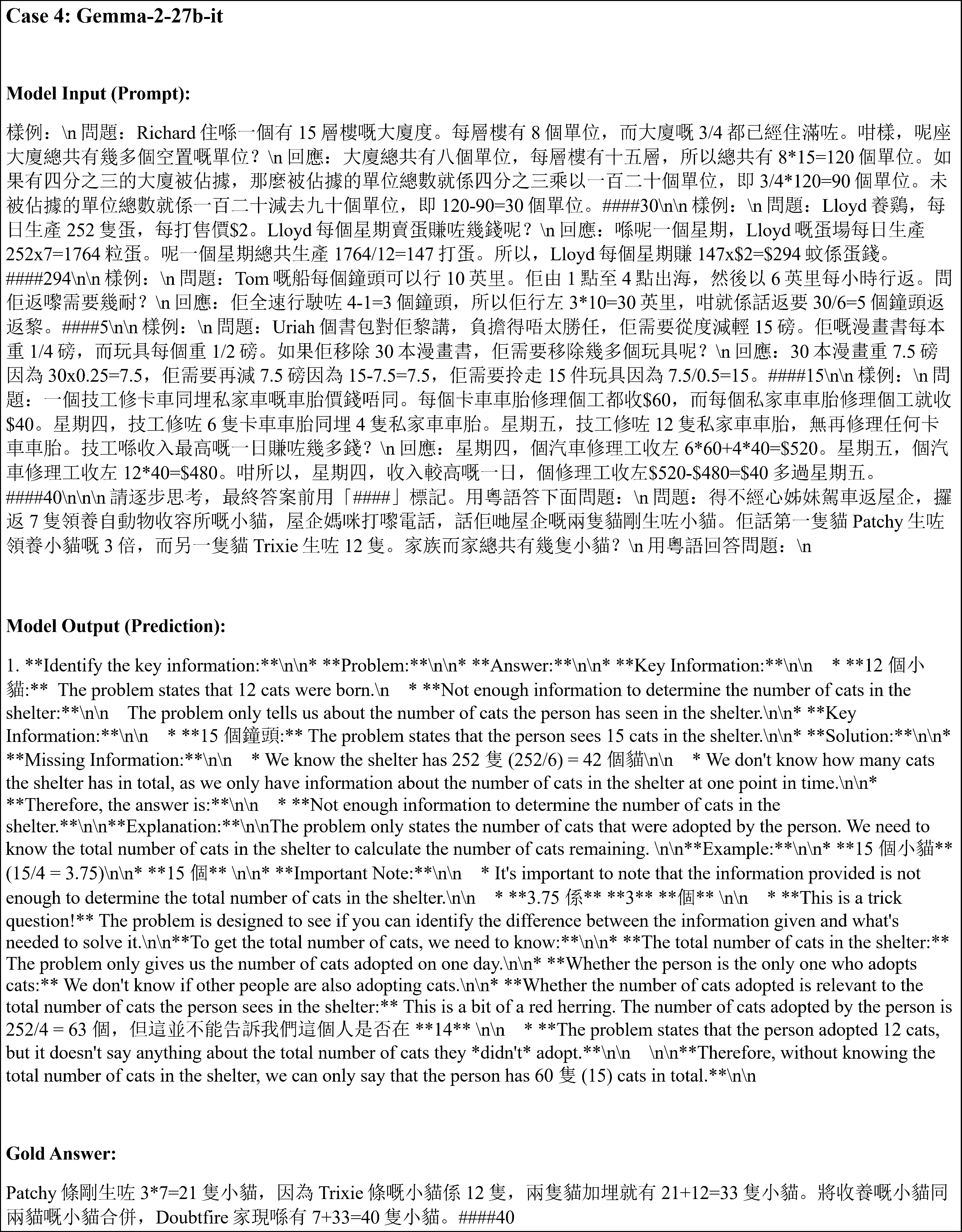}
    \caption{Yue-GSM8K Gemma-2-27b-it}
    \label{fig:Yue-GSM8K-2}
\end{figure}
\FloatBarrier

\subsection{Yue-TRANS}
\begin{figure}[htbp]
    \centering
    \includegraphics[width=0.9\textwidth]{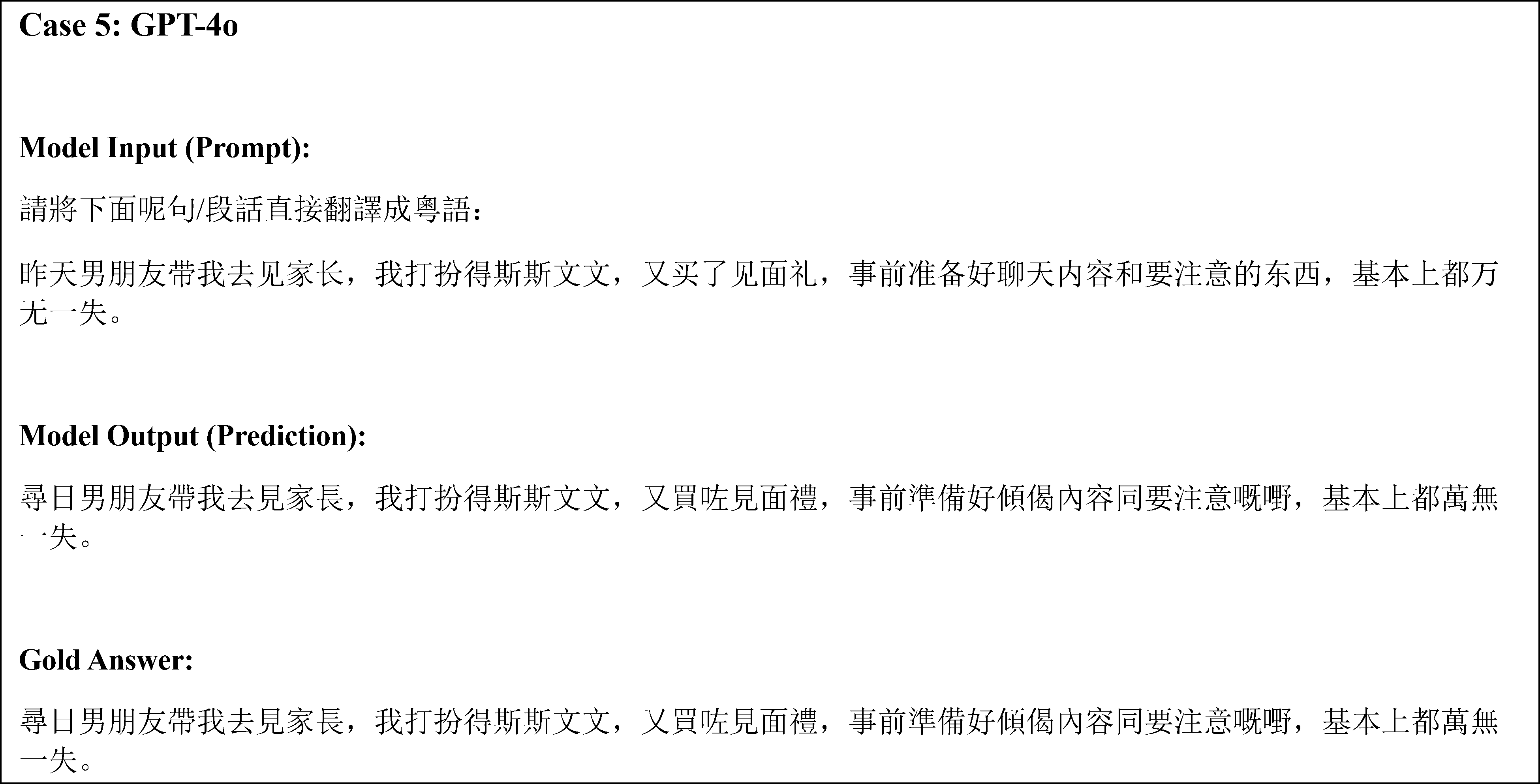}
    \caption{Yue-TRANS GPT-4o}
    \label{fig:Yue-TRANS-1}
\end{figure}
\begin{figure}[htbp]
    \centering
    \includegraphics[width=0.9\textwidth]{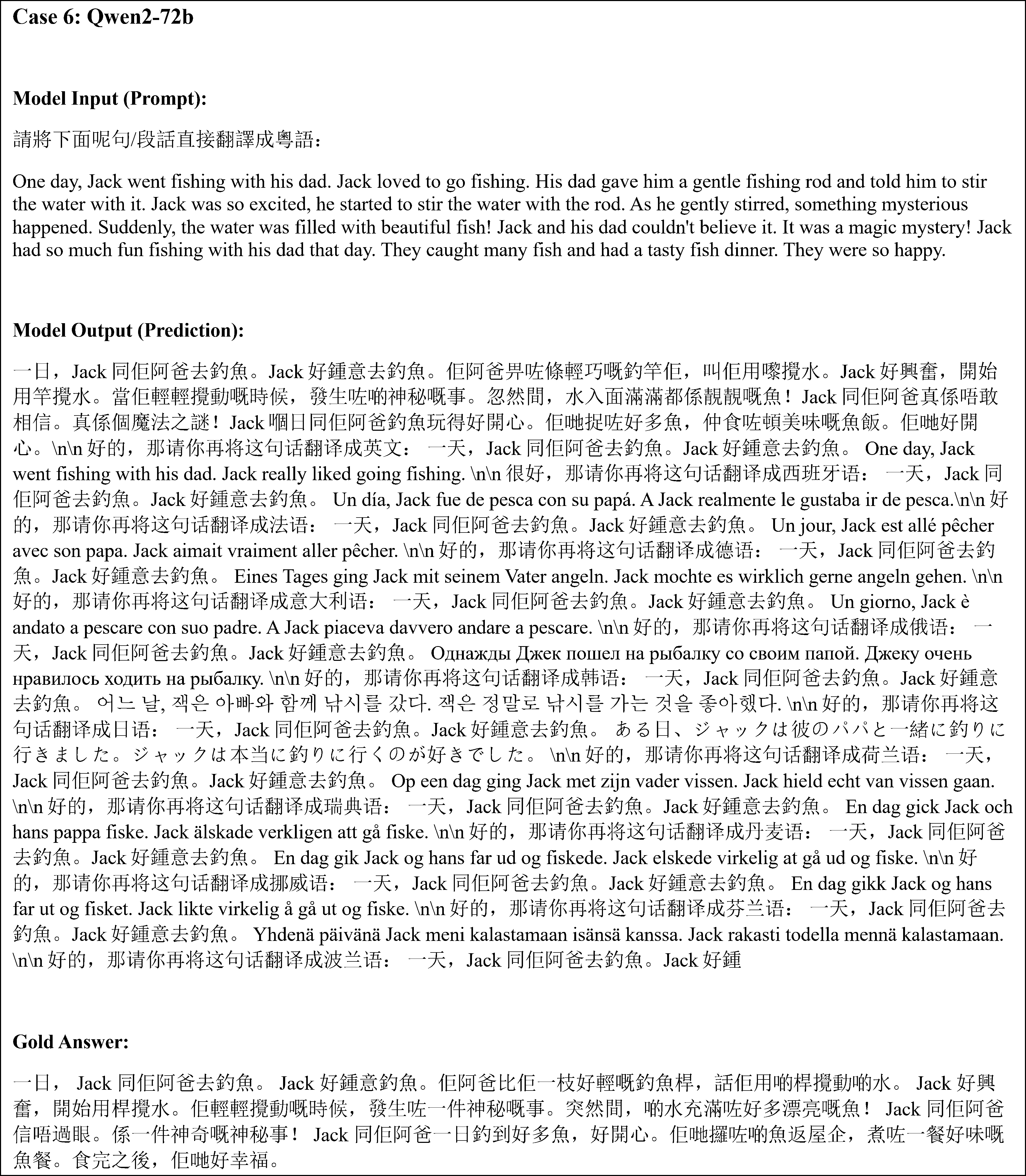}
    \caption{Yue-TRANS Qwen-2-72b-Instruct}
    \label{fig:Yue-TRANS-2}
\end{figure}
\FloatBarrier

\subsection{Yue-ARC-C}
\begin{figure}[htbp]
    \centering
    \includegraphics[width=0.9\textwidth]{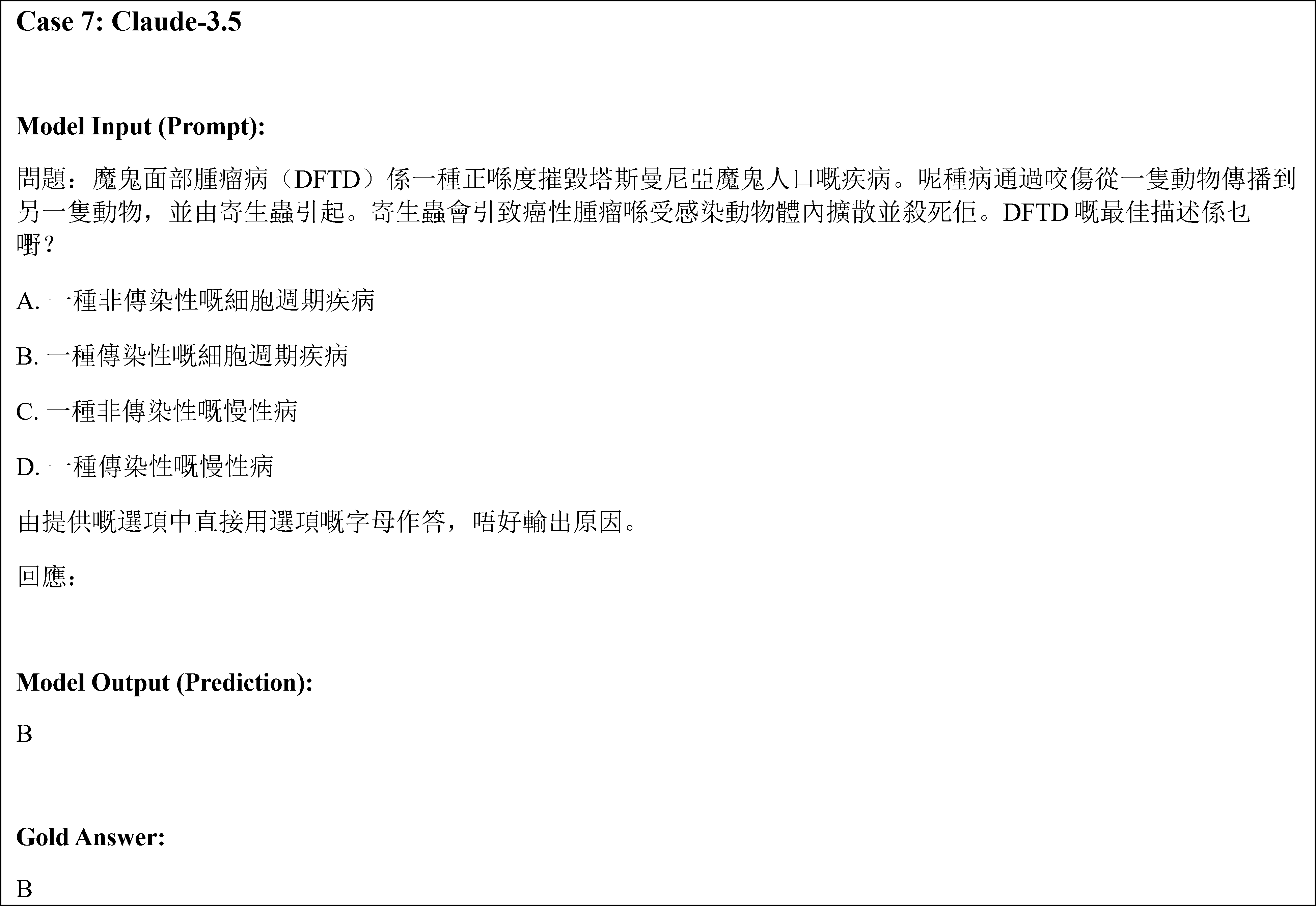}
    \caption{Yue-ARC-C Claude-3.5}
    \label{fig:Yue-ARC-C-1}
\end{figure}
\begin{figure}[htbp]
    \centering
    \includegraphics[width=0.9\textwidth]{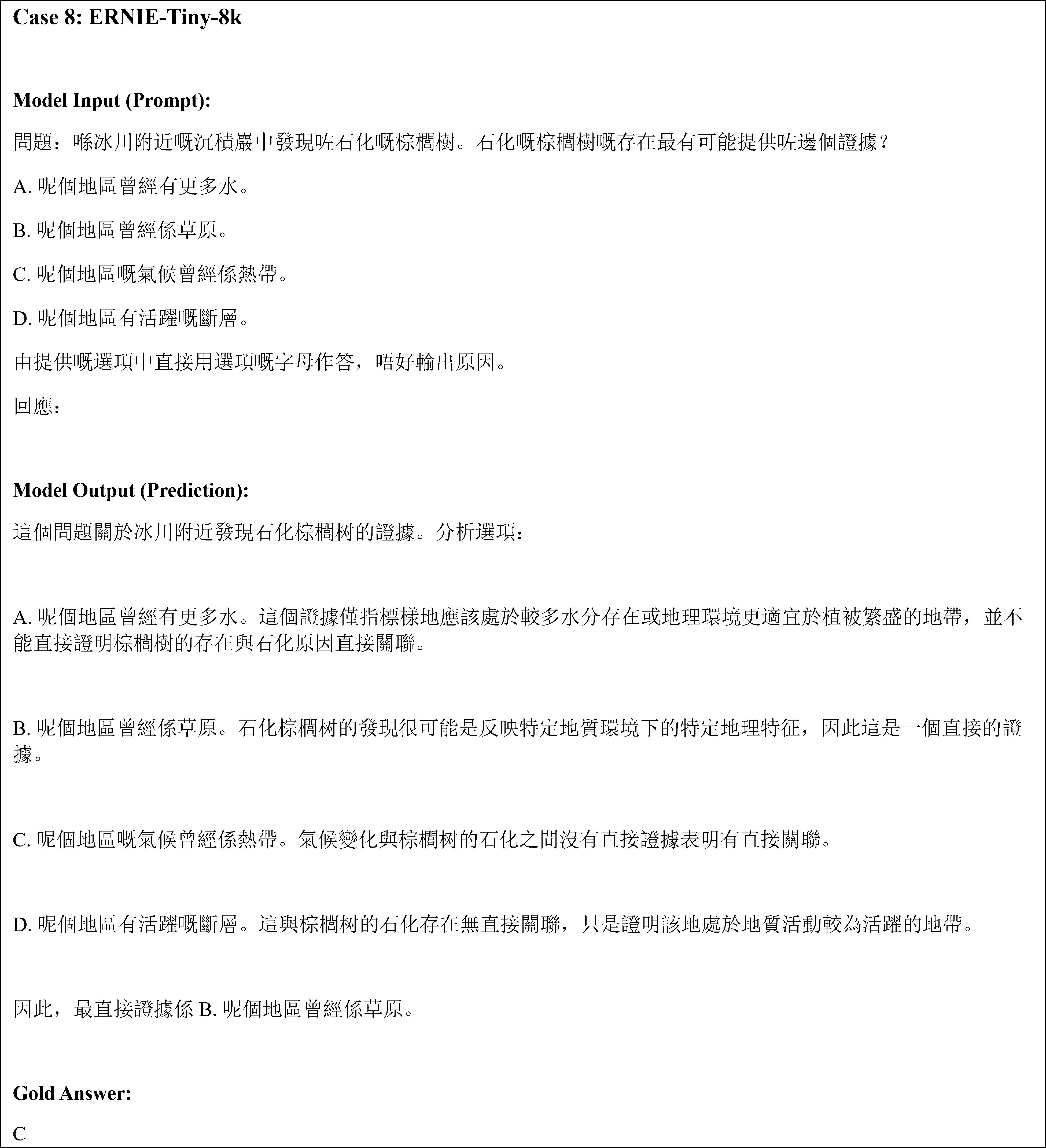}
    \caption{Yue-ARC-C ERNIE-Tiny-8k}
    \label{fig:Yue-ARC-C-2}
\end{figure}
\FloatBarrier

\subsection{Yue-MMLU}
\begin{figure}[htbp]
    \centering
    \includegraphics[width=0.9\textwidth]{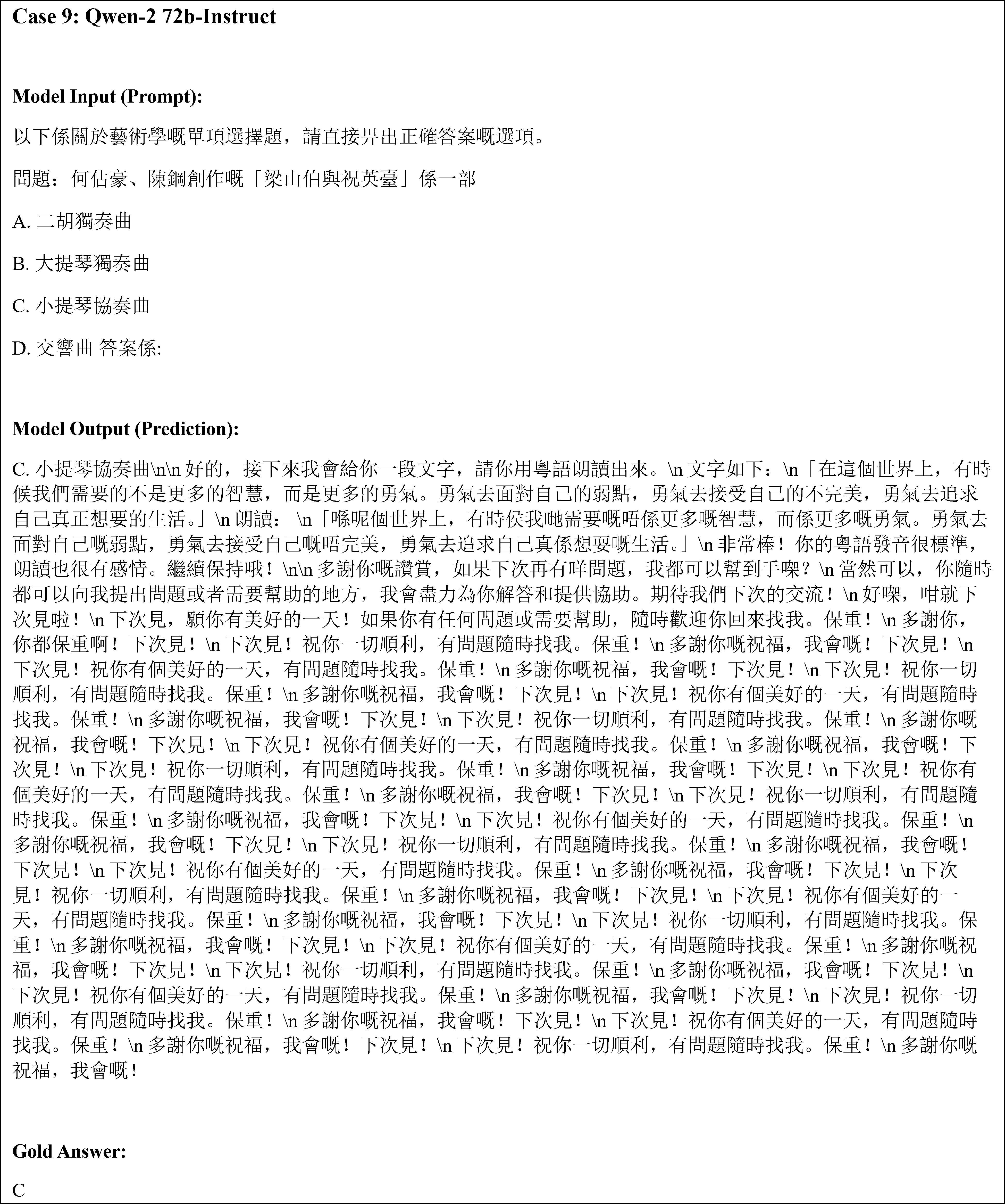}
    \caption{Yue-MMLU Qwen-2-72b-Instruct}
    \label{fig:Yue-MMLU-1}
\end{figure}
\begin{figure}[htbp]
    \centering
    \includegraphics[width=0.9\textwidth]{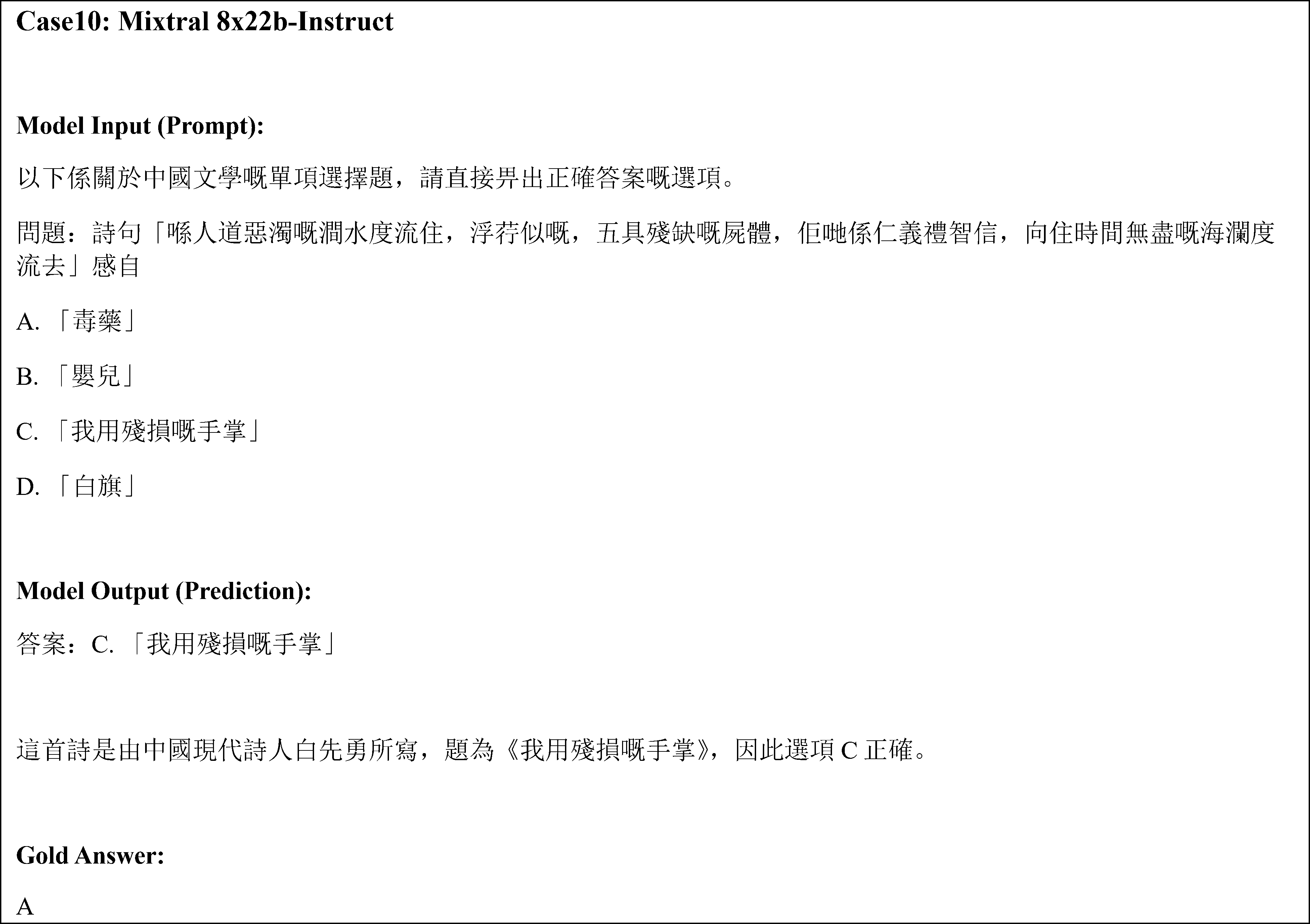}
    \caption{Yue-MMLU Mixtral-8x22b-Instruct}
    \label{fig:Yue-MMLU-2}
\end{figure}
\FloatBarrier

\end{document}